\documentclass[10pt,twocolumn,letterpaper]{article}

 \usepackage[pagenumbers]{cvpr} 

\usepackage[dvipsnames]{xcolor}

\definecolor{cvprblue}{rgb}{0.21,0.49,0.74}
\usepackage[pagebackref,breaklinks,colorlinks,citecolor=cvprblue]{hyperref}
\usepackage{multicol,multirow}   

\usepackage{threeparttable}
\hypersetup{
 colorlinks=true,
 linkcolor=red,
 citecolor=cyan,
 filecolor=magenta, 
 urlcolor=cyan,
 } 
\usepackage{pifont}
 \newcommand{\cmark}{\ding{51}}%
\let\oldding\ding
\renewcommand{\ding}[2][1]{\scalebox{#1}{\oldding{#2}}}%
\def\paperID{12} 
\def\confName{CVPR}
\def\confYear{2024}

\title{HiPA: Enabling One-Step Text-to-Image Diffusion Models\\via High-Frequency-Promoting Adaptation}

\author{Yifan Zhang\\
National University of Singapore\\ 
{\tt\small yifan.zhang@u.nus.edu}
\and
Bryan Hooi\\
National University of Singapore\\ 
{\tt\small bhooi@comp.nus.edu.sg}
}

\begin{document}
\maketitle
\begin{abstract}
Diffusion models have revolutionized text-to-image generation, but their real-world applications are hampered by the extensive time needed for hundreds of diffusion steps. Although progressive distillation has been proposed to speed up diffusion sampling to 2-8 steps, it still falls short in one-step generation, and necessitates training multiple student models, which is highly parameter-extensive and time-consuming. To overcome these limitations, we introduce High-frequency-Promoting Adaptation (HiPA), a parameter-efficient approach to enable one-step text-to-image diffusion. Grounded in the insight that high-frequency information is essential but highly lacking in one-step diffusion, HiPA focuses on training one-step, low-rank adaptors to specifically enhance the under-represented high-frequency abilities of advanced diffusion models. The learned adaptors empower these diffusion models to generate high-quality images in just a single step. Compared with progressive distillation, HiPA achieves much better performance in one-step text-to-image generation (37.3$\rightarrow$23.8 in FID-5k on MS-COCO 2017) and 28.6x training speed-up (108.8$\rightarrow$3.8 A100 GPU days), requiring only 0.04\% training parameters (7,740 million $\rightarrow$ 3.3 million). We also demonstrate HiPA's effectiveness in text-guided image editing, inpainting and super-resolution tasks, where our adapted models consistently deliver high-quality outputs in just one diffusion step. The source code will be released. 
 
\end{abstract}
    
\section{Introduction}
\label{sec:intro}

Text-to-image generation~\cite{rombach2022high, balaji2022ediffi, saharia2022photorealistic,ruiz2023dreambooth}, aiming at synthesizing images from textual descriptions, has undergone a significant transformation with the advent of diffusion models~\cite{song2020score,dhariwal2021diffusion,ho2020denoising,zhang2023adding}. These models, known for their multi-step denoising process, have set new benchmarks in the quality of generated images, marked by increased fidelity and detail~\cite{Betker2023Improving}. 
However, the necessity for multiple diffusion steps — each meticulously refining the image — results in significantly long generation time. This diminishes the practicality of text-to-image diffusion models for real-time applications, adversely affecting user experience~\cite{salimans2021progressive,song2023consistency,geng2023one,li2023autodiffusion}.

\begin{figure}[t]
 \vspace{0.15in}
	\centering
	\includegraphics[width=1\linewidth]{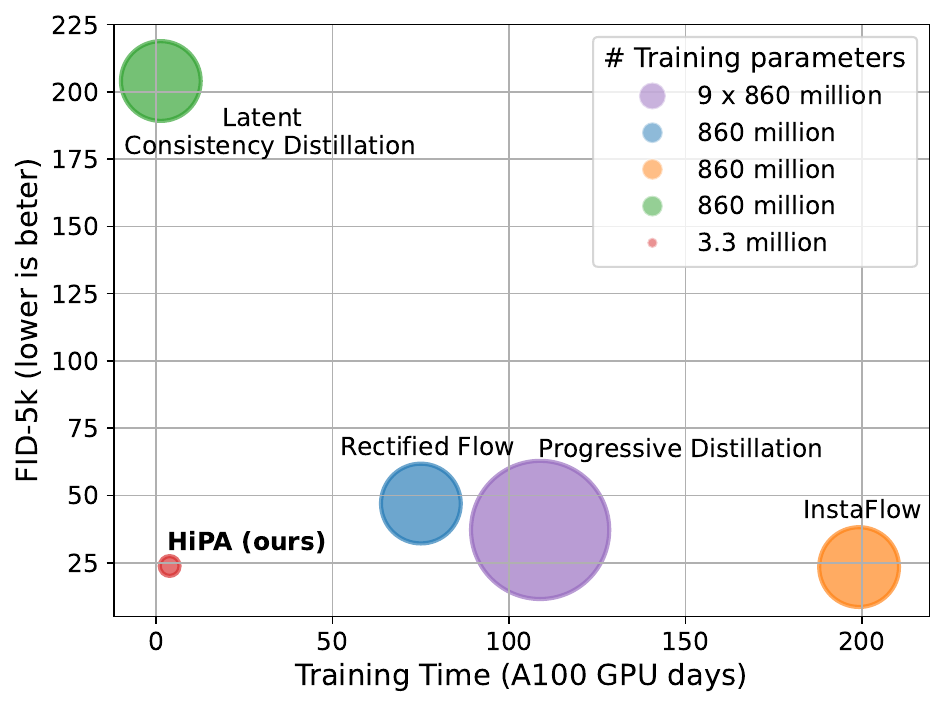}
 \vspace{-0.15in}
	\caption{Performance of one-step text-to-image diffusion on MS-COCO 2017~\cite{lin2014microsoft}. We observe that our HiPA performs remarkably well in terms of FID while requiring much less computation time and fewer training parameters.}
	\label{figure1} 
 \vspace{-0.1in}
\end{figure}

\begin{figure*}[t] 
	\centering 
	\includegraphics[width=0.9\linewidth]{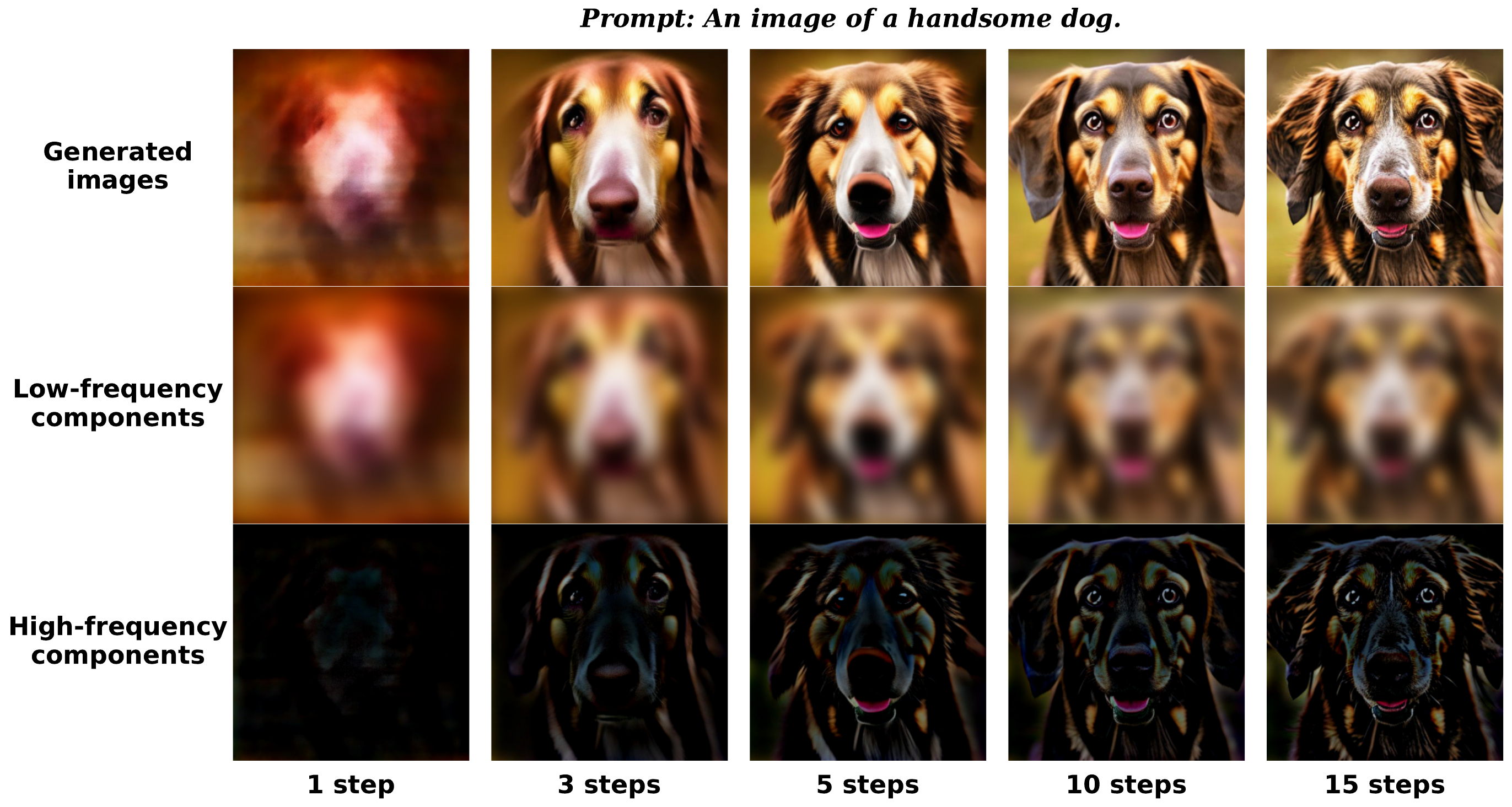}
 
	\caption{Illustration of text-to-image generation with different diffusion steps based on Stable Diffusion~\cite{rombach2022high} and DPM sampler~\cite{lu2022dpm}. Initially, simple low-frequency components form, followed by complex high-frequency details that increase realism. Notably, one-step diffusion images lack the complex high-frequency components, making them noticeably less realistic.} 
	\label{figure2} 
\end{figure*}

To mitigate this issue, progressive distillation (PD)~\cite{salimans2021progressive,meng2023distillation,li2023snapfusion,berthelot2023tract} proposes to distill a $T$-step teacher diffusion model into a new $T/2$-step student model, and repeats this process until fewer-step diffusion models are achieved. Despite enabling more efficient diffusion over 2-8 steps, PD strategies usually require conducting distillation multiple times for training multiple student models, leading to a highly slow and parameter-heavy process. Although consistency distillation~\cite{song2023consistency,song2023improved} can accelerate unguided diffusion models, its effectiveness in text-to-image generation remains unverified. As an extension, latent consistency distillation~\cite{luo2023latent} facilitates 2-4 step text-to-image generation, but still falls short in one-step generation.

In this work, we focus on one-step generation for streamlining the inference efficiency of text-to-image diffusion models. Instead of conducting slow progressive distillation by training multiple student models with extensive parameters, we propose a parameter-efficient High-frequency-Promoting Adaptation (HiPA) approach to 
accelerate existing advanced multi-step models to one-step diffusion. This acceleration is achieved by training low-rank HiPA adaptors, which contain significantly fewer parameters compared to PD. While low-rank adaptation~\cite{hu2021lora} has been used to tailor diffusion models for particular tasks~\cite{gu2023mix,smith2023continual}, how to train the adaptors for efficient one-step diffusion acceleration remains an unresolved challenge.
 
To figure this challenge out, we delve into the multi-step generation process of text-to-image diffusion models, aiming to uncover what information one-step diffusion lacks, compared to its multi-step counterpart. As shown in Figure~\ref{figure2}, we identify a pivotal nuance in the text-to-image generation process — diffusion begins with generating low-frequency information, followed by producing high-frequency details with the increase of diffusion steps. It is worth noting that one-step diffusion often struggles to produce rich high-frequency details, which, however, are essential for realistic image generation.  Existing acceleration techniques, such as progressive distillation~\cite{meng2023distillation,li2023snapfusion} and consistency models~\cite{song2023consistency,song2023improved}, overlook this crucial aspect, thus sacrificing high-frequency detail generation in one-step diffusion and leading to limited image quality.

In light of these findings, HiPA trains the low-rank adaptor to specifically boost high-frequency detail generation in one-step diffusion. Central to HiPA is a new diffusion adaptation loss, consisting of a spatial perceptual loss and a high-frequency promoted loss. The spatial perceptual loss ensures the structural coherence in the generated images, while the high-frequency promoted loss, leveraging Fourier transform and edge detection, is specifically designed to enhance the subtle, yet crucial, high-frequency details. This dual-loss strategy effectively preserves detailed textures and edges that are often overlooked in one-step diffusion, facilitating rapid generation without significantly compromising image quality.

Our approach is rigorously validated through extensive experiments in one-step text-to-image generation, demonstrating that HiPA outperforms existing one-step methods in both visual fidelity and training efficiency, while requiring much fewer training parameters. As illustrated in Figure~\ref{figure1} and Table~\ref{table_training_efficiency}, compared to progressive distillation, HiPA significantly improves one-step text-to-image generation performance (37.3 to 23.8 in FID-5k on MS-COCO 2017), accelerates training by 28.6 times (108.8 to 3.8 A100 GPU days), and drastically reduces training parameter needs (7,740 million to just 3.3 million). To showcase HiPA's versatility, we extend its application to text-guided image editing, inpainting, and super-resolution tasks, where we reduce the number of diffusion steps to a single step. Promising results demonstrate HiPA's remarkable potential for efficient and practical use in various real-world image modification and enhancement applications.


\begin{figure*}[t] 
 \vspace{-0.25in}
 \centerline{\includegraphics[width=13.5cm]{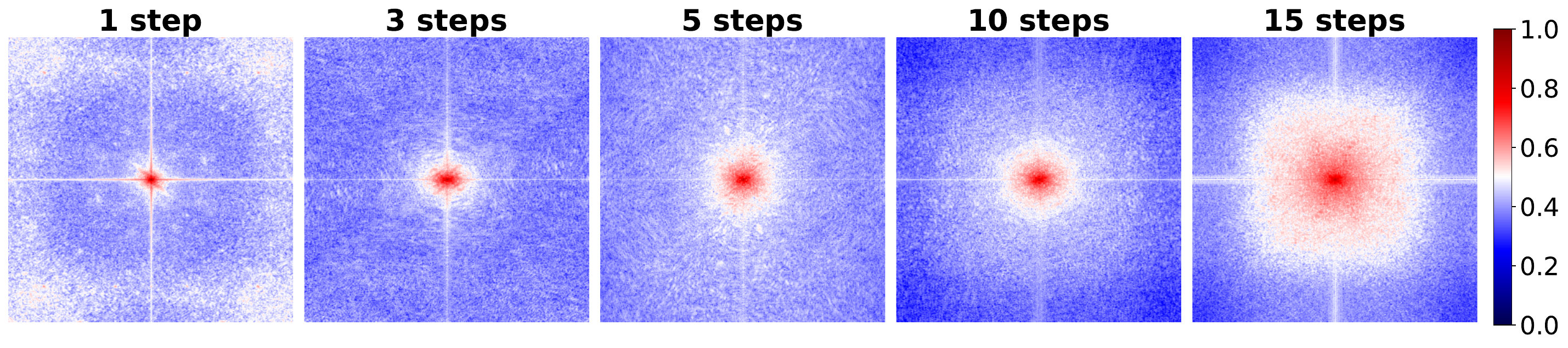}} \vspace{-0.15in}
 \caption{Power Spectral Density analysis of the generated images by Stable Diffusion with different diffusion steps (DPM sampler).}\label{figure3} 
 \vspace{-0.1in}
\end{figure*}
 
\section{Related Work}
\label{sec:related} 
\noindent \textbf{Text-to-image generation.} Image generation~\cite{azizi2023synthetic,dai2023disentangling,sariyildiz2023fake,tian2023stablerep,zhou2023training,zhang2023expanding,liu2023geom} has evolved through paradigms including Generative Adversarial Networks~\cite{esser2021taming,isola2017image,cao2019multi}, auto-regressive models~\cite{kingma2019introduction,ramesh2021zero}, and notably, diffusion models~\cite{dhariwal2021diffusion,ho2020denoising}. 
In the realm of text-to-image generation, diffusion models have emerged as powerful tools for text-guided, high-fidelity image synthesis. Transformative models like DALL-E~\cite{ramesh2021zero,ramesh2022hierarchical,Betker2023Improving}, Imagen~\cite{saharia2022photorealistic}, and Stable Diffusion~\cite{rombach2022high}, have demonstrated remarkable zero-shot generation capabilities. However, the developments in diffusion models have also led to challenges in balancing generation efficiency and effectiveness, since they often require many diffusion steps to generate images with high quality. For example, Stable Diffusion~\cite{rombach2022high}, even with advanced samplers like DDIM~\cite{song2020denoising} and DPM~\cite{lu2022dpm}, typically requires more than 15-50 steps to generate high-quality images.

\noindent \textbf{Acceleration of text-to-image diffusion.} Strategies for accelerating diffusion models fall into two main categories. Initially, researchers proposed fast post-hoc samplers~\cite{song2020denoising,lu2022dpm,Zhang2023dpm,zheng2023fast,dockhorn2022genie} to decrease the number of inference steps to between 15 and 50. However, these enhancements at inference were not enough, which prompts a new paradigm of model adaptation~\cite{yang2023diffusion,ruiz2023dreambooth}. One advancement is progressive distillation~\cite{salimans2021progressive,meng2023distillation,li2023snapfusion,berthelot2023tract}, which distills pre-trained diffusion models to fewer than 10 steps~\cite{rombach2022high}. Despite offering improvements for 2-8 step diffusion, these methods often necessitate repeated distillation of multiple student models, resulting in a slow and parameter-heavy process. Although consistency distillation~\cite{song2023consistency,song2023improved} and equilibrium models~\cite{geng2023one} offer acceleration for unguided diffusion models, their applicability to text-to-image generation is yet unproven. As an extension, latent consistency distillation~\cite{luo2023latent} works for 2-4 step text-to-image generation, but it is still limited in one-step generation. Instaflow~\cite{liu2023insta} enables one-step text-to-image generation by training the whole stable diffusion through 2-stage reflow and 2-stage distillation, which require over 199 A100 GPU days. Distinct from these methods, our HiPA approach is more parameter-efficient by only training the low-rank HiPA adaptors for acceleration (around 3.8 A100 GPU days), thereby circumventing the extensive time and computational resources required for progressive distillation. By deliberately promoting high-frequency generation abilities, our HiPA effectively enables one-step text-to-image diffusion models.

\noindent \textbf{Low-rank adaptation} (LoRA)~\cite{hu2021lora} is a parameter-efficient strategy for diffusion model customization~\cite{gu2023mix,smith2023continual}.
Given a weight matrix $W\in \mathbb{R}^{d \times d}$, LoRA adapts it by introducing two \emph{low-rank adaptors}: $W' = W + UV^T$, where $U \in \mathbb{R}^{d \times k}$ and $V \in \mathbb{R}^{d \times k}$, and $k \ll d$. The product $UV^T$ signifies a low-rank update to the frozen weight matrix, constraining the adaptation within a subspace and significantly cutting down the number of parameters involved. This requires updates to only $O(kd)$ parameters, a stark reduction compared to the original $O(d^2)$, making the model adaptation notably more manageable and resource-efficient.

\begin{figure*}[t] 
 \vspace{-0.2in}
 \centerline{\includegraphics[width=14.8cm]{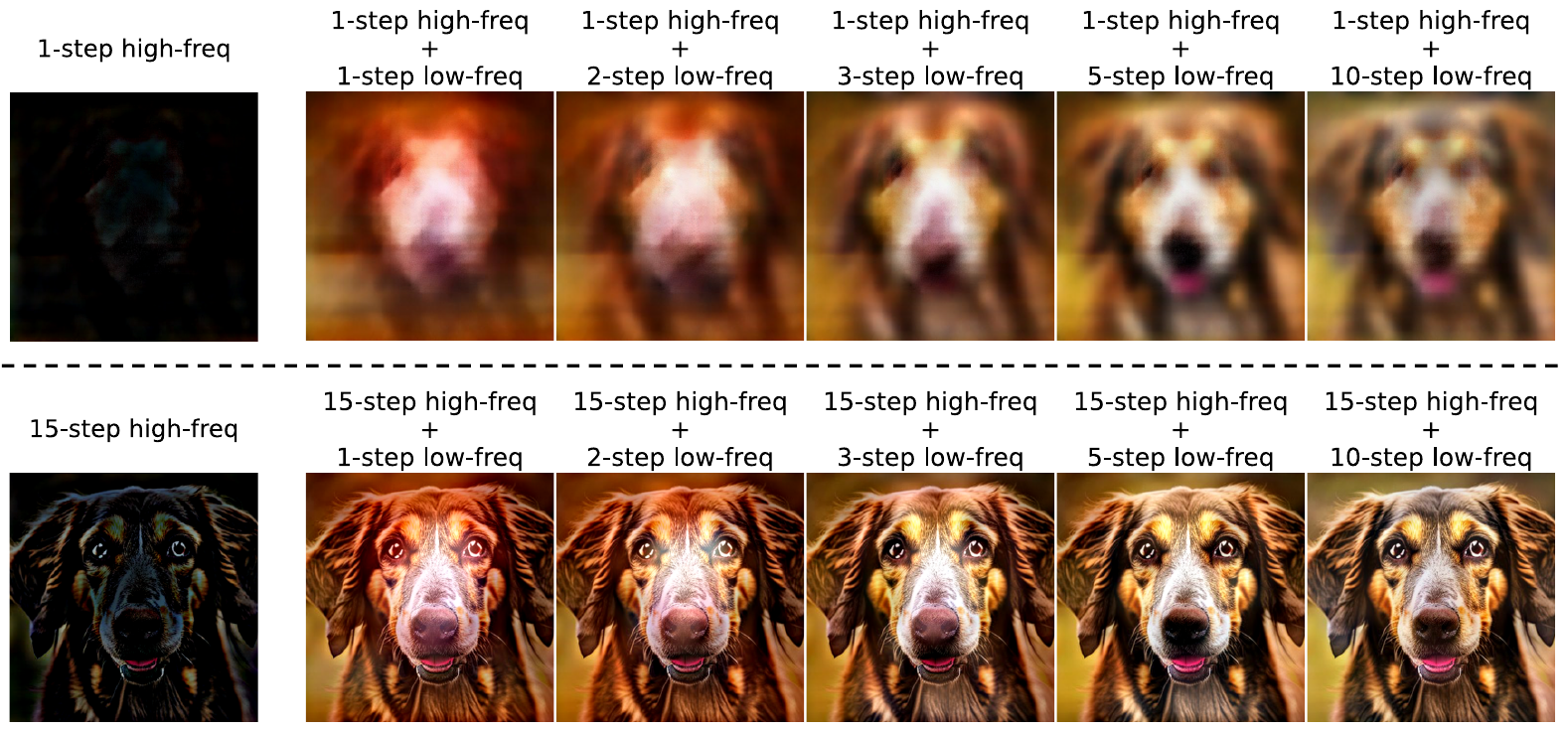}} 
 \vspace{-0.05in}
 \caption{Illustration of the impact of high-frequency components in enhancing image clarity for one-step text-to-image diffusion. Combining the high-frequency components from the 15-step images with the the low-frequency components from fewer-step images results in sharper images after Inverse Fourier Transform, while using one-step high-frequency components provides no clarity enhancement.}\label{figure_cross} 
\end{figure*}

\section{Preliminary Studies}
\label{sec:preliminary}
\subsection{One-step diffusion lacks high-frequency details}
 
To advance one-step text-to-image diffusion, we begin with an analysis of Stable Diffusion (SD)~\cite{rombach2022high}, aiming to dissect the nuances of images produced at different diffusion steps. As shown in Figure~\ref{figure2} (first row), we identify an essential characteristic of text-to-image generation: the images generated by one-step SD are notably blurry, and their quality dramatically improves with an increase in diffusion steps. To delve into this phenomenon, we leverage Discrete Fourier Transform~\cite{gonzales1987digital} to differentiate between high and low-frequency information within the image, and then employ Inverse Fourier Transform~\cite{gonzales1987digital} to reconstruct images for visualization. Figure~\ref{figure2} (second row) shows that the generation process initially focuses on generating foundational elements and the underlying scene of the image, such as general color schemes and brightness. Interestingly, as the number of diffusion steps increases, the initial foundational elements remain relatively stable. Meanwhile, the generated images progressively incorporate more intricate details, including object edges, complex textures, and distinctive patterns (cf. Figure~\ref{figure2}, last row). 

To substantiate this observation, we use Power Spectral Density analysis~\cite{bovik2010handbook} to shed light on the intricacies of this progression. As shown in Figure~\ref{figure3}, there is a discernible pattern where diffusion models first extract low-frequency information (\ie, central regions of the spectrum), subsequently enriching the image with high-frequency details (\ie, peripheral regions of the spectrum). This further corroborates our observation.
To summarize, Figures~\ref{figure2}-\ref{figure3} highlight a stark insufficiency of high-frequency components in one-step generated images, which explains the inferior performance of SD in a single step. 

\subsection{High frequency matters for one-step diffusion}

To delve deeper into the critical role of high-frequency information in few-step generation, we conduct an experiment involving a mixup of high and low-frequency components, followed by visualizing the reconstructed images. Specifically, we use Discrete Fourier Transform to extract high and low-frequency components from images generated through 1 to 15 steps. We then cross-combine the high-frequency components from the 1 and 15-step images with the low-frequency components from the 1, 2, 3, 5, and 10-step images. Afterwards, we use Inverse Fourier Transform to visualize the images resulting from each combination.

As shown in Figure~\ref{figure_cross}, we find that combining the high-frequency components from the 15-step image with the low-frequency components from fewer-step images yields images of noticeably enhanced clarity and quality. In contrast, when the high-frequency components are sourced from the 1-step image, the resulting images continue to exhibit their intrinsic blurriness, regardless of the low-frequency source. This observation is crucial, as it not only emphasizes the pivotal role of high-frequency components in text-to-image diffusion, but also shows that enhancing one-step diffusion models with superior high-frequency generation distinctly improves the quality of one-step synthetic images.

\subsection{Promoting high-frequency generation boosts one-step diffusion models}

Building upon the above observations, we hypothesize that enhancing the high-frequency generation could boost one-step text-to-image diffusion models. To validate this, we evaluate the influence of high-frequency promotion on low-rank adaptation of SD for one-step generation.
Specifically, we train the low-rank adaptors by aligning the one-step generated images of the adapted SD with those from the original SD but with more diffusion steps (\eg, 10 or 15 DPM steps). We explore the alignment via several adaptation losses, including L2 spatial loss, spatial loss augmented with low-frequency promotion, and spatial loss enhanced with high-frequency promotion. Here, high-frequency promotion is realized by aligning the Fourier-reconstructed high-frequency images between one-step and multi-step counterparts using L2 loss. Low-frequency promotion is implemented in a similar way.

The results, as shown in Table~\ref{table_preliminary}, reveal a notable trend. The use of just an L2 spatial loss yields a noticeable improvement compared to the SD baseline. However, when this loss is augmented with low-frequency promotion, there is a counterproductive effect, leading to diminished performance. This suggests that prioritizing low-frequency information may actually hinder one-step diffusion. Crucially, the incorporation of high-frequency promotion alongside L2 spatial loss results in the most substantial gains, achieving superior realism (lower FID), higher diversity (higher IS), and better textual fidelity (higher CLIP score) in one-step generated images. These results compellingly verify our insight that promoting high-frequency abilities in diffusion model adaptation significantly enhances one-step text-to-image generation performance. 

\begin{table}[t] 
 \begin{center}
 \begin{threeparttable} 
 \resizebox{0.47\textwidth}{!}{
 	\begin{tabular}{lccc}\toprule 
 Methods & FID-30k $\downarrow$ & IS $\uparrow$ & CLIP $\uparrow$ \cr
 \midrule
 Stable Diffusion & 355.2	 	& 2.0 & 0.11	 \\
 ~~$\bullet$~L2 spatial loss & 131.2	 	& 8.3 & 0.16 \\
 ~~$\bullet$~L2 spatial loss + low-frequency promotion & 163.5 & 4.7	 &	 0.15 \\
 ~~$\bullet$~L2 spatial loss + high-frequency promotion & \textbf{115.2}	& \textbf{10.6}	& \textbf{0.18} \\ 
 \bottomrule 
	\end{tabular}}
	 \end{threeparttable}
	 \end{center} 
 \vspace{-0.15in}
	\caption{One-step generation performance of Stable Diffusion adaptation on MS-COCO 2014~\cite{lin2014microsoft}.}
	\label{table_preliminary} 
 \vspace{-0.1in}
\end{table} 
\section{Our Approach}
\label{sec:method}

In light of the aforementioned insights, we propose a new parameter-efficient strategy, High-frequency-Promoting Adaptation (HiPA), to enable one-step text-to-image diffusion models.

\begin{figure}[t]
\vspace{-0.2in}
	\centering
	\includegraphics[width=0.7\linewidth]{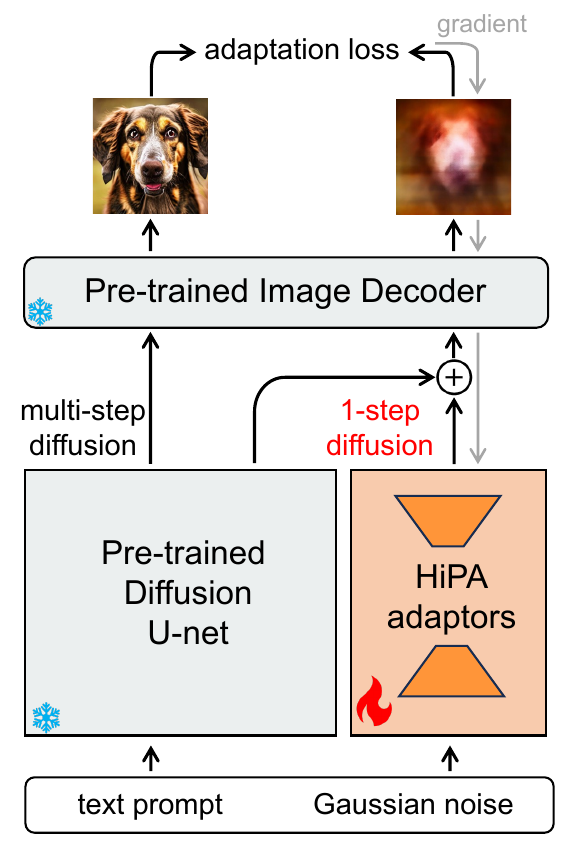}
\vspace{-0.1in}
	\caption{An illustration of our parameter-efficient High-frequency-Promoting Adaptation (HiPA) approach.}
	\label{framework}
\vspace{-0.1in}
 
\end{figure}

\vspace{0.1in}
\noindent \textbf{Overall scheme.} 
Our approach diverges from previous methods like Progressive Distillation~\cite{meng2023distillation} and InstaFlow~\cite{liu2023insta}, which focus on tuning the entire pre-trained diffusion models. Instead, our approach aims to train a low-rank HiPA adaptor to enhance the one-step generation abilities of diffusion models.
As shown in Figure~\ref{framework}, HiPA achieves this by aligning the images generated in a single step by the adapted model with those produced by the original, frozen model across multiple steps (\eg, 15 DPM steps). To specifically promote high-frequency abilities, HiPA employs a composite adaptation loss. This loss integrates a spatial perceptual loss with a high-frequency promoted loss, collectively refining the model's one-step generation for improved fidelity and high-frequency details:
\begin{equation}
L_{\text{adaptation}} = L_{\text{spatial}} + L_{\text{high-freq}}.
\end{equation}

\vspace{0.05in}
\noindent \textbf{Spatial perceptual loss $L_{\text{spatial}}$.}
We compute the spatial adaptation loss based on the Deep Image Structure and Texture Similarity (DISTS) metric \cite{ding2020image}:
\begin{equation}
L_{\text{spatial}} = L_{\text{DISTS}}\left(I_{\text{generated}}^{\text{1-step}}, I_{\text{generated}}^{\text{multi}}\right),
\end{equation}
where $I_{\text{generated}}^{\text{1-step}}$ and $I_{\text{generated}}^{\text{multi-step}}$ denote the images generated by the HiPA-adapted model in a single step and those by the original diffusion model with multiple steps. DISTS goes beyond pixel-level differences to capture perceptual dissimilarities between images, considering both structural and textural characteristics. We find it empirically outperforms the L2 and LPIPS~\cite{zhang2018unreasonable} in our paradigm.

\begin{figure}[t] 
	\centering
	\includegraphics[width=0.9\linewidth]{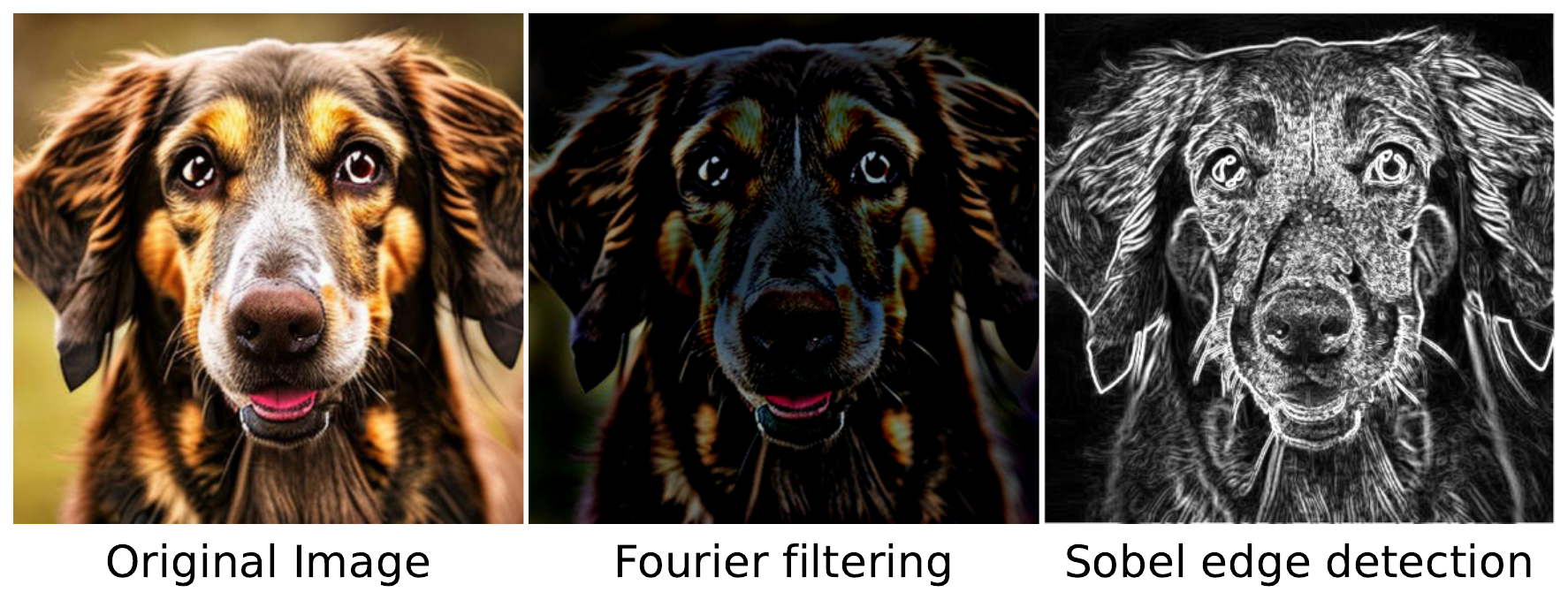}\vspace{-0.1in}
	\caption{Illustration of the extracted high-frequency information.}
	\label{high-freq}
\vspace{-0.1in} 
\end{figure}

\vspace{0.1in}
\noindent \textbf{High-frequency promoted loss $L_{\text{high-freq}}$.}
To effectively promote high-frequency abilities, we apply two complementary strategies to extract high-frequency information: Fourier transform and edge detection. As illustrated in Figure~\ref{high-freq}, the two approaches complement each other in highlighting different aspects of high-frequency details.
 
For the Fourier strategy, we first apply Discrete Fourier Transform (DFT)~\cite{gonzales1987digital} to the generated image $I_{\text{generated}}$, transforming it from the spatial to the frequency domain. The high-frequency components are then extracted through high-pass filtering, followed by Inverse Fourier Transform (IFT)\cite{gonzales1987digital} to reconstruct the high-frequency image $I_{\text{freq}}$. This process can be described by:
\begin{align} 
I_{\text{freq}} &= \text{IFT}\Big{(}\text{DFT}(I_{\text{generated}}) \odot M_{\text{high}}(u, v)\Big{)},
\end{align} 
where $M_{\text{high}}(u, v)$ is a high-pass filter in frequency domain.

\begin{figure*}
 \vspace{-0.15in}
 \centering 
 \begin{minipage}{.32\textwidth}
 \centering
 \includegraphics[width=\linewidth]{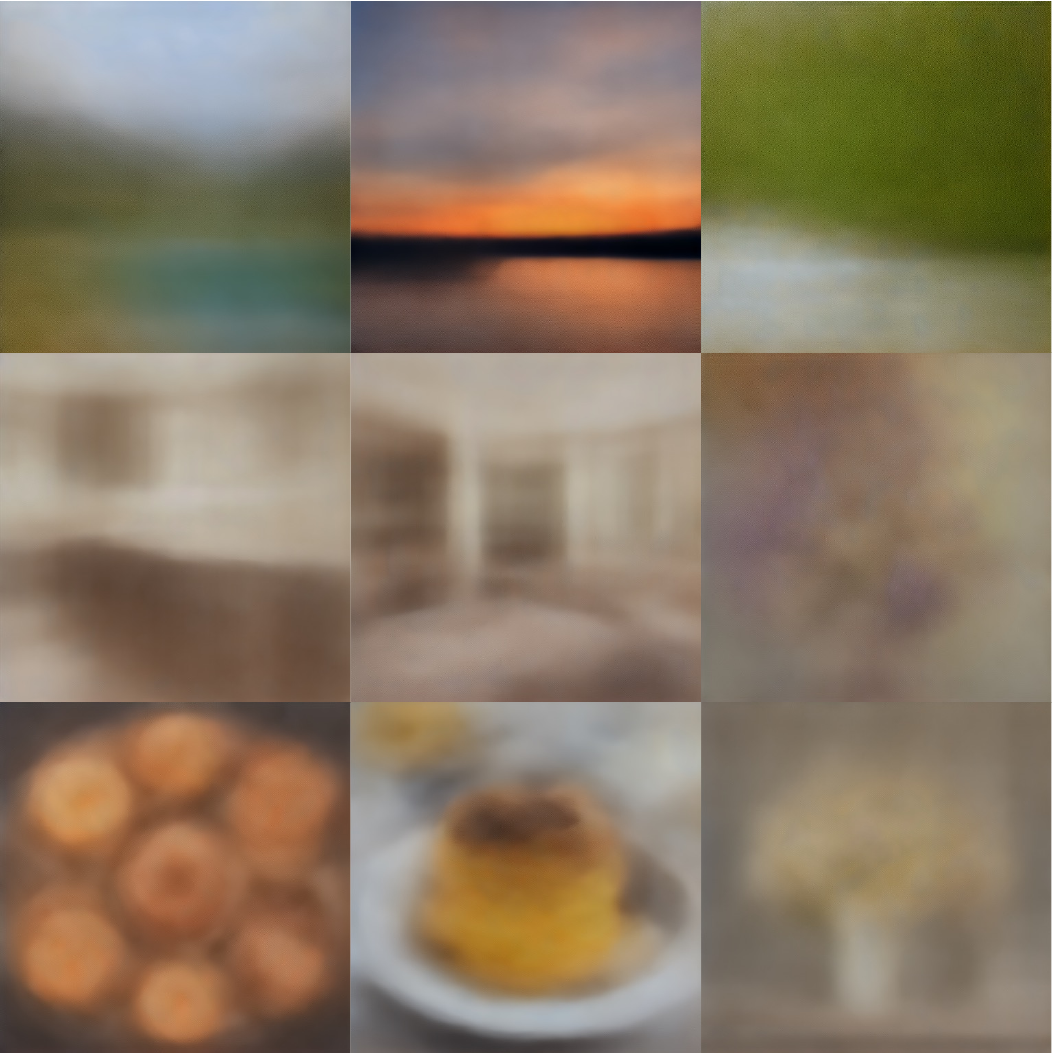}
 \vspace{-0.25in}
 \caption*{SD~\cite{rombach2022high}} 
 \end{minipage}
 \hfill 
 \begin{minipage}{.32\textwidth}
 \centering
 \includegraphics[width=\linewidth]{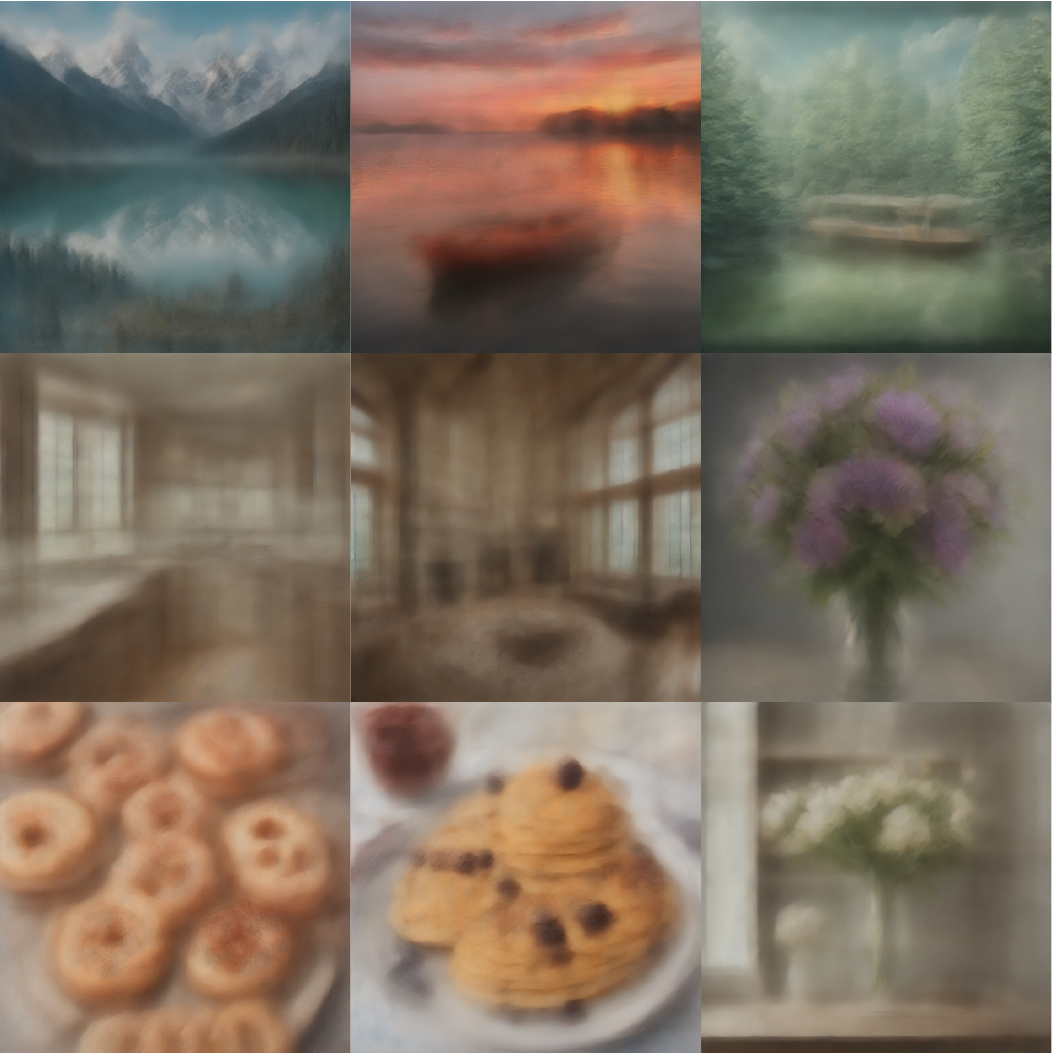}
 \vspace{-0.25in}
 \caption*{Latent Consistency~\cite{luo2023latent}} 
 \end{minipage}
 \hfill
 \begin{minipage}{.32\textwidth}
 \centering
 \includegraphics[width=\linewidth]{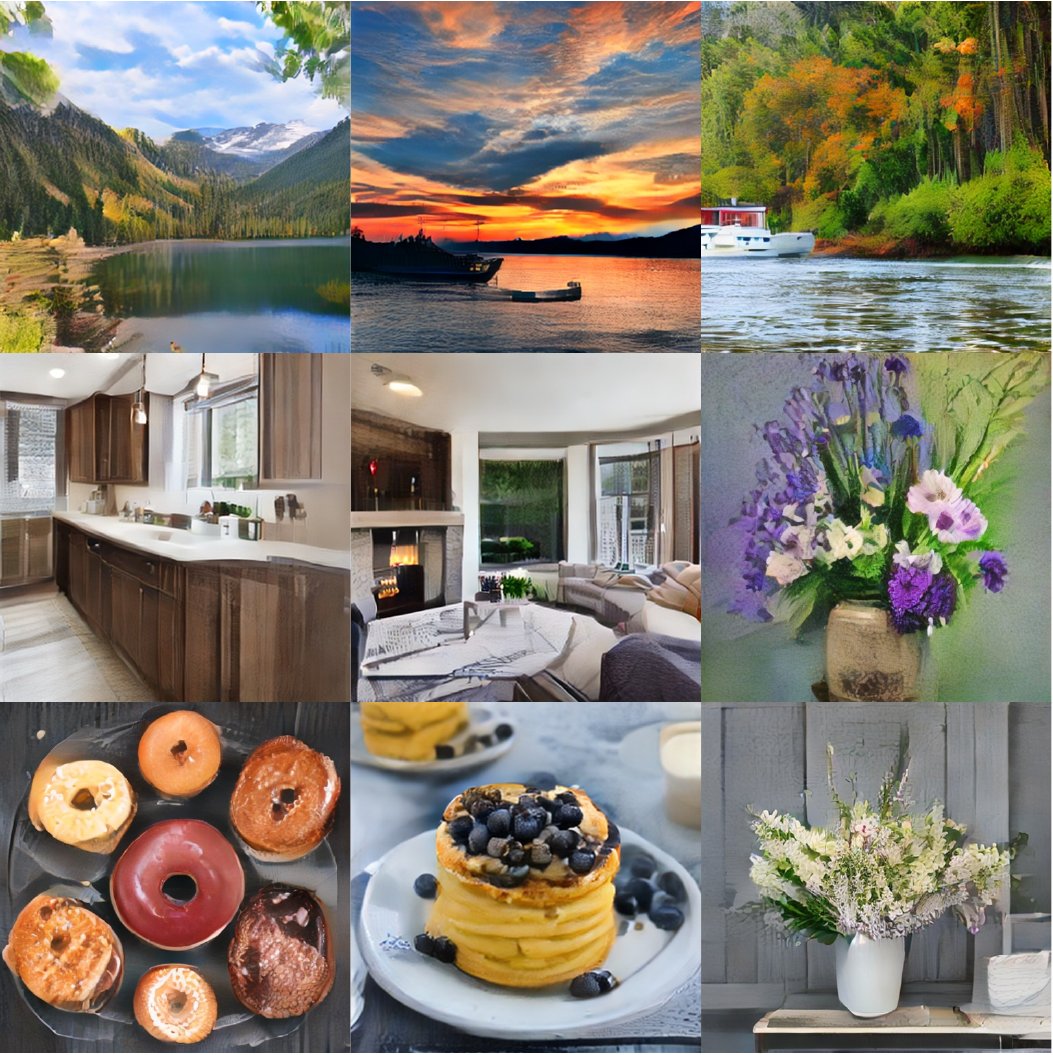}
 \vspace{-0.25in}
 \caption*{HiPA-adapted SD (ours)} 
 \end{minipage} 
 \vspace{-0.1in}
 \caption{One-step text-guided image generation on MS-COCO 2014 ($512\times512$) by Stable Diffusion (SD), Latent Consistency Distillation, and HiPA. These visual results show that our approach can generate high-quality images in a single diffusion step.}
\label{visualization_qualitative}
 \vspace{-0.1in}
\end{figure*}

Meanwhile, we apply the Sobel operator~\cite{castleman1996digital} to extract edge information of images. This operator computes the image gradient $\nabla I$ through convolution of $I$ with pre-defined horizontal and vertical kernels $G_x$ and $G_y$, thereby highlighting significant intensity transitions. The detected edge image $I_{\text{edge}}$ is computed by:
\begin{equation}
I_{\text{edge}} = \sqrt{ (I_{\text{generated}} * G_x)^2 + (I_{\text{generated}} * G_y)^2 },
\end{equation}
where $*$ represents the convolution operation, and the Sobel kernels are defined as $G_x = [-1, 0, 1; -2, 0, 2; -1, 0, 1]$ and $G_y = [-1, -2, -1; 0, 0, 0; 1, 2, 1]$.

Based on the extracted high-frequency information, we design the high-frequency promoted loss by aligning the high-frequency details of the one-step images with those of the multi-step generated images:
\begin{equation}
L_{\text{high-freq}} = L_{\text{DISTS}}\left(I_{\text{freq}}^{\text{1-step}}, I_{\text{freq}}^{\text{multi}}\right)+L_{\text{DISTS}}\left(I_{\text{edge}}^{\text{1-step}}, I_{\text{edge}}^{\text{multi}}\right).
\end{equation}
In this way, the learned HiPA adaptor enables the adapted one-step model to emulate the superior quality of its multi-step counterpart, particularly in generating high-frequency details that are pivotal for image realism and quality.
 
\section{Experiments}
\label{sec:experiments}

In this section, we evaluate the effectiveness and versatility of our method in one-step text-to-image diffusion. We begin with the experimental settings.

\begin{table}[t] 
 \begin{center}
 \begin{threeparttable} 
 \resizebox{0.47\textwidth}{!}{
 	\begin{tabular}{lccccc}\toprule 
 Methods& Step & FID-30k $\downarrow$ & IS $\uparrow$ & CLIP $\uparrow$ & Inference time \cr
 \midrule
 Stable Diffusion~\cite{rombach2022high} & 25 & 9.40 	& 31.83 & 0.29	 & 8.6 hours \\
 \midrule
 Stable Diffusion~\cite{rombach2022high} & 1 & 355.21	 	& 1.97 & 0.11 & 
 2.3 hours \\ 
 Latent Consistency~\cite{luo2023latent} & 1 & 195.35 & 4.46 & 0.20 & 3.1 hours \\ 
 HiPA (ours) & 1 & 13.91 & 28.09 	& 0.31 & 2.5 hours \\ 
 \bottomrule 
	\end{tabular}}
	 \end{threeparttable}
	 \end{center} 
 \vspace{-0.15in}
	\caption{One-step generation performance on COCO 2014. The guidance scale is 2. Inference time for 30k image generation is measured in A100 GPU hours.} 
	\label{table_main_result} 
 \vspace{-0.15in}
\end{table}

\subsection{Experimental setups}

\noindent \textbf{Datasets and Metrics.} We principally use the MS-COCO 2017 training set~\cite{lin2014microsoft} for diffusion model adaptation, and use the COCO 2014/2017 validation set for evaluation. This dataset offers diverse textual-visual content, making it an ideal benchmark for method evaluation. Moreover, we use three main evaluation metrics: Fréchet Inception Distance (FID), Inception Score (IS), and CLIP score (ViT-g/14).

\vspace{0.1in}
\noindent \textbf{Implementation details.} HiPA is developed using the Diffusers library~\cite{diffusers}. Default settings include an adaptor rank of $r=16$ and a learning rate of $1e-4$. The high-pass filter's cutoff for high-frequency extraction is set to 5. To improve image smoothness, we also regularize the pixel variation~\cite{rudin1992nonlinear} of the generated images. DPM-solver~\cite{Zhang2023dpm} is used for both training and inference.
Training, constrained by GPU capacity, uses a batch size of 8 on NVIDIA RTX 3090 GPUs; while for a fair comparison of computational costs, we record training and inference time using an A100 GPU. The ablation studies on hyper-parameters (\eg, learning rate, adaptor rank, cutoff) can be found in Appendices. Moreover, we implement Stable Diffusion~\cite{rombach2022high} and Latent Consistency~\cite{luo2023latent} via their official code on Diffusers.

\begin{table}[t] 
 \begin{center}
 \begin{threeparttable} 
 \resizebox{0.47\textwidth}{!}{
 	\begin{tabular}{lccccc}\toprule 
 Methods& Step & FID-5k $\downarrow$ & Training time & \# Params. of training \cr 
 \midrule
 Stable Diffusion~\cite{rombach2022high} & 8 	 & 31.7$^*$ & 6,250 Days$^{\dag}$ &	 	 860 million \\
 SnapFusion~\cite{li2023snapfusion} & 8 & 	 	 24.2$^*$ & - & 3$\times$848 million \\
 \midrule 
 Progressive Distillation~\cite{meng2023distillation} & 1 & 	 37.2$^{\dag}$ & 108.8 Days$^{\dag}$ & 9$\times$860 million \\
 
 2-Rectified Flow~\cite{liu2023insta} & 1 & 	 47.0$^{\dag}$ & 75.2 Days$^{\dag}$ & 860 million\\ 
 InstaFlow-0.9B~\cite{liu2023insta} & 1 & 	 23.4$^{\dag}$ & 199.2 Days$^{\dag}$ &	 860 million\\ 
 Latent Consistency~\cite{luo2023latent} & 1 & 	 204.0 & 1.3 Days$^{\ddag}$ 	& 860 million\\ 
 HiPA (ours) & 1 & 23.8 & 3.8 Days	& 3.3 million \\ 
 \bottomrule 
	\end{tabular}}
	 \end{threeparttable}
	 \end{center} 
 \vspace{-0.15in}
	\caption{Generation performance and training costs on COCO 2017. Training time is measured in A100 GPU days. Symbol $*$ indicates the values from SnapFusion~\cite{li2023snapfusion}, ${\dag}$ from InstaFlow~\cite{liu2023insta}, and ${\ddag}$ from Latent Consistency~\cite{luo2023latent}. Note: the parameter numbers for SnapFusion and Progressive Distillation take into account the distillation of multiple student models (\ie, 3 and 9, respectively).}
 \vspace{-0.1in}
	\label{table_training_efficiency} 
\end{table} 
 
\begin{figure*}[t] \vspace{-0.2in}
	\centering
	\includegraphics[width=1\linewidth]{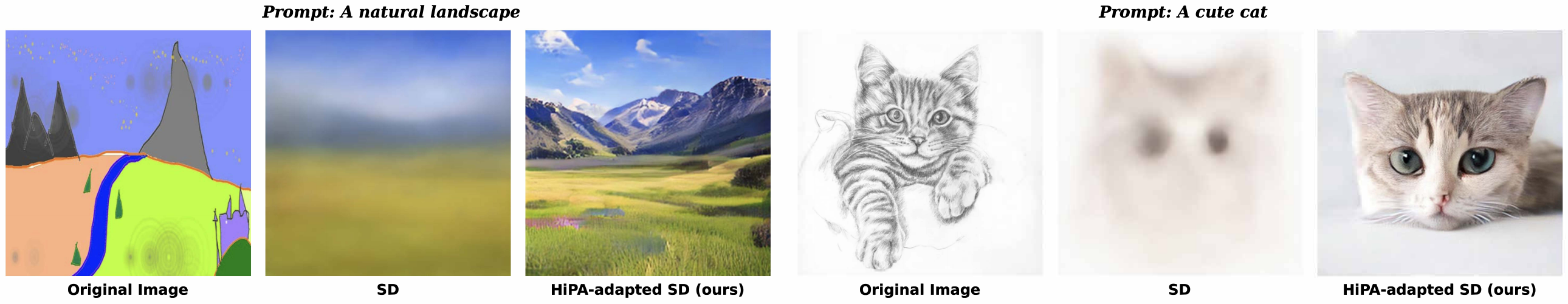}
 \vspace{-0.25in}
	\caption{One-step text-guided image editing by the original Stable Diffusion (SD) and our HiPA-adapted SD model. More visualization examples are available in the supplementary material.}
 \vspace{-0.1in}
	\label{visualization_image2image} 
\end{figure*}

\begin{figure*}[t]
	\centering
	\includegraphics[width=1\linewidth]{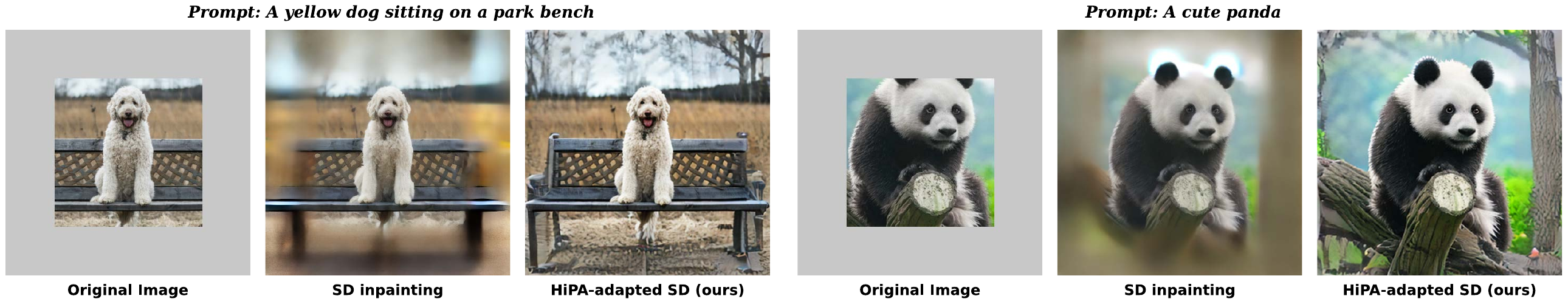}
 \vspace{-0.25in}
	\caption{One-step text-guided image inpainting by Stable Diffusion (SD) inpainting model and our HiPA-adapted SD inpainting model. More visualizations are provided in the supplementary material.}
	\label{visualization_inpainting} 
 
 \vspace{-0.15in}
\end{figure*}

\begin{figure*}[t]
\vspace{-0.15in}
\centering
    \includegraphics[width=1\linewidth]{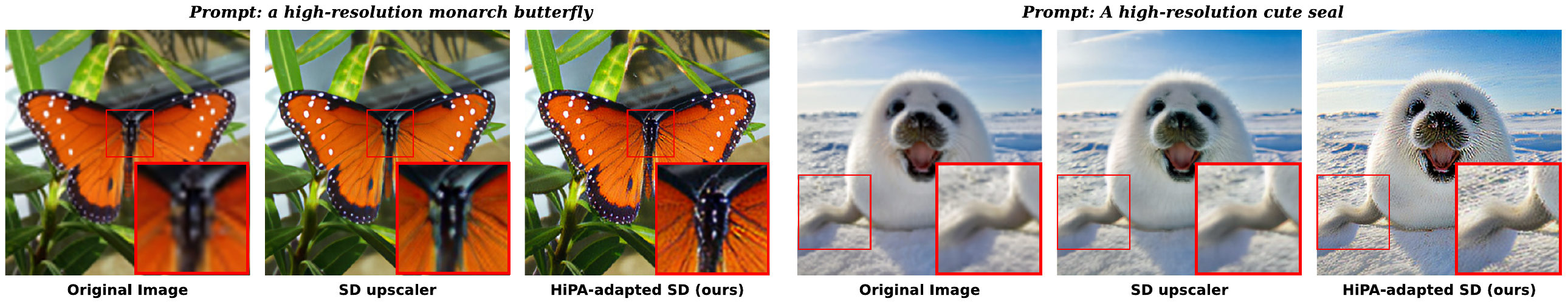}
    \vspace{-0.25in}
    \caption{One-step text-guided image super-resolution by Stable Diffusion (SD) upscaler model and our HiPA-adapted upscaler model.  More visualized results can be found in the supplementary material.} 
\vspace{-0.1in}
\label{visualization_sr} 
\end{figure*}

\subsection{One-step text-guided image generation}\label{text2image}
 
In this work, we focus on adapting the Stable Diffusion (SD) v2.1 base model~\cite{rombach2022high} for rapidly generating high-fidelity images from text descriptions. Our method trains the adaptors for 5 epochs on the COCO 2017 training set, followed by qualitative and quantitative assessments.

The one-step generated images, as showcased in Figure~\ref{visualization_qualitative}, demonstrate a remarkable improvement in quality over both one-step SD and Latent Consistency~\cite{luo2023latent}. This is backed by the quantitative results in Tables~\ref{table_main_result}-\ref{table_training_efficiency}, where HiPA outperforms existing methods across FID, IS, and CLIP. These improvements are not just numerical; they translate into perceptibly more realistic images (lower FID), greater diversity (higher IS), and better alignment with textual descriptions (higher CLIP). Importantly, HiPA enables one-step text-to-image diffusion, offering a considerable boost in inference efficiency over the original multi-step SD. This advancement is attained with only a slight reduction in performance, effectively balancing speed and quality, making it highly suitable for real-world applications.

A more critical superiority of HiPA lies in its training efficiency (cf.~Table~\ref{table_training_efficiency}). Unlike Progressive Distillation that necessitates training multiple student models (up to 9 for one-step generation), HiPA solely trains low-rank adaptors. This not only reduces the training time to a mere 3.8 A100 GPU days, compared to Progressive Distillation's extensive 108.8 days, but also significantly cuts down on the number of training parameters—from hundreds of millions in other models to just 3.3 million in HiPA. These advantages position HiPA as an effective, resource-efficient solution, setting a new standard in one-step text-to-image diffusion.

\subsection{One-step text-guided image editing} 
We proceed with experiments on one-step text-guided image editing based on SDEdit~\cite{meng2022sdedit}. Following SDEdit, we first perturb the input image with Gaussian noise in the latent space, and then apply the original or our HiPA-adapted SD models for one-step diffusion. As shown in Figure~\ref{visualization_image2image}, our adapted model impressively yields high-quality style-transfer images in just a single step, demonstrating a noticeable improvement over the one-step image editing capabilities of the standard SD.

\subsection{One-step text-guided image inpainting} 
We next apply our approach to accelerate latent inpainting diffusion, based on the open-source SD v2 inpainting model\footnote{\url{https://huggingface.co/stabilityai/stable-diffusion-2-inpainting}}. This
model is a fine-tuned variant of SD, enhanced to handle masks and masked images with additional input channels. Specifically, we apply HiPA to adapt the inpainting model on COCO 2017 for 1 epoch, where we retain the central image content while masking out 50\% of the peripheral pixels. We train the HiPA adaptors to effectively inpaint these masked regions in just one diffusion step, aligning the output with that of the original inpainting model's 15-step diffusion. The qualitative results, visualized in Figure~\ref{visualization_inpainting}, highlight our method's capability to facilitate fast, effective text-guided image inpainting for real-world applications.

\subsection{One-step text-guided super-resolution}
We further extend HiPA to accelerate latent super-resolution diffusion, based on the widely available SD 4x-upscaler model\footnote{\url{https://huggingface.co/stabilityai/stable-diffusion-x4-upscaler}}. This model, a specialized variant of SD, is tailored for text-guided super-resolution. We adapt this model on COCO 2017 for 2,000 iterations, beginning with images downsampled to $128\times 128$. We train the HiPA adaptors to upscale these images to $512\times512$ in a single diffusion step by aligning their outputs with those from the original model's 15-step diffusion. The visualized results, as shown in Figure~\ref{visualization_sr}, highlight the capacity of HiPA to perform fast text-guided image super-resolution in real scenarios.

\subsection{Discussions}

\paragraph{Ablation studies on our adaptation loss.} 

We explore the impact of different losses within HiPA on the quality of one-step generation. The results, presented in Table~\ref{table_loss_ablation}, demonstrate that incorporating the high-frequency promoted loss enhances image quality. Notably, the combination of high-frequency Fourier loss and high-frequency edge loss shows a complementary effect, further elevating image quality. This validates the effectiveness of our adaptation loss in improving one-step text-to-image diffusion.

\begin{table}[t] 
 \begin{center}
 \begin{threeparttable} 
 \resizebox{0.47\textwidth}{!}{
 	\begin{tabular}{cccccc}\toprule 
 \multicolumn{3}{c}{Losses within HiPA} && \multirow{2}{*}{FID-30k $\downarrow$} & \multirow{2}{*}{CLIP $\uparrow$} \\\cmidrule{1-3}
 Spatial loss & Fourier high-frequency & Edge high-frequency & & & \cr
 
 \midrule
 \cmark && && 17.57 	& 0.29	 \\
 \cmark & \cmark & && 17.02 &	 0.30 \\
 \cmark && \cmark &&16.15 & 0.30 \\
 \cmark &\cmark&\cmark && 15.89 & 0.31 \\ 
 \bottomrule 
	\end{tabular}}
	 \end{threeparttable}
	 \end{center} 
 \vspace{-0.15in}
	\caption{Ablation studies of losses within HiPA, \ie, spatial loss and high-frequency losses (including Fourier transform one and edge detection one) on MS-COCO 2014. For fast evaluation, all methods are trained for only 1 epoch.}
 \vspace{-0.1in}
	\label{table_loss_ablation} 
\end{table} 
 
\begin{figure}[t] 
 \vspace{-0.05in}
 \centering
 \includegraphics[width=0.8\linewidth]{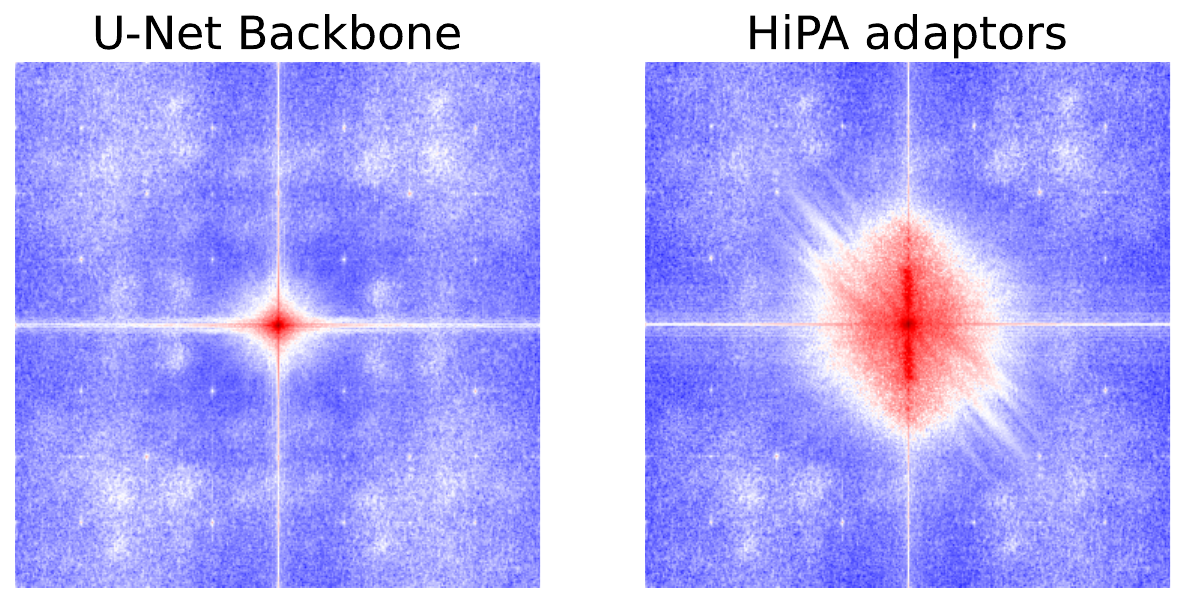} 
 \vspace{-0.1in}
 \caption{Power Spectral Density analysis for the generated features by U-Net backbone and the low-rank HiPA adaptors.} 
 \label{figure_lora}
\end{figure}

\paragraph{What information does the learned adaptor generate?} 
In probing the unique contributions of low-rank adaptors, we use Power Spectral Density analysis~\cite{bovik2010handbook} to dissect the features obtained from different components of the HiPA-adapted model. As depicted in Figure~\ref{figure_lora}, the backbone primarily contributes low-frequency information (\ie, central regions of the spectrum), whereas the learned HiPA adaptor enhances high-frequency information (\ie, peripheral regions of the spectrum). This demonstrates the effectiveness of HiPA in promoting the high-frequency generation abilities of one-step text-to-image diffusion models. 
 
\vspace{0.1in}
\noindent \textbf{Safety check.} In the realm of image generation, ethical considerations are of utmost importance. To ensure responsible use, we conduct safety checks on images generated by our adapted model using the Google Cloud Vision API\footnote{\url{https://cloud.google.com/vision/docs/detecting-safe-search}}, a deep learning-powered tool designed to analyze image content. As shown in Table~\ref{safety_check}, our model's synthetic images are verified as safe and non-harmful, with the majority being classified as either "Very unlikely" or "Unlikely" to contain questionable content across all evaluated criteria. The visual examples in Figures~\ref{visualization_qualitative}-\ref{visualization_sr} also corroborate the safe and appropriate nature of our generated images.

\vspace{0.1in}
\noindent \textbf{Limitations and Solutions.} Our exploration reveals a trade-off between speed and quality in text-to-image diffusion models. As shown in the supplementary, there are still occasional artifacts in our one-step text-guided generated images. However, many of these issues may stem from the original multi-step SD models, evident in distortions like faces and fingers. Future enhancements could include adapting more advanced diffusion models (\eg SD-XL~\cite{podell2023sdxl} and DALL-E3~\cite{shi2020improving}), dependent on the access to GPUs with greater memory capacity. Despite this limitation, our adapted model serves as an effective tool for quick image previews. Once users select their preferred draft images, they can switch to the original SD model without the HiPA adaptors for generating higher-quality images, thus balancing between quick drafts and detailed final images. 
 
\begin{table}[t] 
 \begin{center}
 \begin{threeparttable} 
 \resizebox{0.45\textwidth}{!}{
 	\begin{tabular}{lccccc} 
 \toprule 
 Metrics & Very unlikely & Unlikely & Neutral & Likely & Very likely \\ \midrule 
 Adult & 85\% & 14\% & 1\% & 0\% & 0\% \\ 
 Spoof & 72\% & 23\% & 3\% & 2\% & 0\% \\ 
 Medical & 86\% & 14\% & 0\% & 0\% & 0\% \\ 
 Violence & 75\% & 25\% & 0\% & 0\% & 0\% \\ 
 Racy & 54\% & 31\% & 11\% & 4\% & 0\% \\
 \bottomrule 
	\end{tabular}}
	 \end{threeparttable}
	 \end{center} 
 \vspace{-0.15in}
	\caption{{Safety check of the generated 30k images by HiPA on COCO 2014, in terms of metrics of Google Cloud Vision API.}} 
 \vspace{-0.1in}
	\label{safety_check} 
\end{table} 
\section{Conclusion}
\label{sec:conclusion} 

To advance one-step text-to-image diffusion models, we have introduced a novel High-frequency-Promoting Adaptation (HiPA) method. HiPA adeptly addresses the computational and qualitative dilemmas posed by existing text-to-image diffusion models. By integrating parameter-efficient adaptation with high-frequency promotion, HiPA not only accelerates the text-to-image diffusion process, but also amplifies the high-frequency details essential for generating photorealistic images. Our empirical evidence underscores HiPA's superiority over conventional methods, demonstrating it as a practical and parameter-efficient approach in real-time text-to-image diffusion generation. 
{   \newpage
    \small
    \bibliographystyle{ieeenat_fullname}
    \bibliography{main}
}

\clearpage
\maketitlesupplementary
\appendix

This supplementary provides more backgrounds in Appendix~\ref{sec:backgrounds}, more details of Fourier loss in Appendix~\ref{sec:rationale}, more experimental and visualized results in Appendices~\ref{app_c}-\ref{app_d}.

\section{Backgrounds}
\label{sec:backgrounds}

\subsection{Diffusion models}
Diffusion models, particularly \textit{Denoising Diffusion Probabilistic Models} (DDPMs), are renowned for generating high-quality images. These models operate by adding and reversing Gaussian noise over a sequence of steps. Formally, starting from \( x_0 \sim p(x_0) \), the data is noised through a chain of transitions:
$p(x_t|x_{t-1}) = \mathcal{N}(x_t; \sqrt{1 - \beta_t} x_{t-1}, \beta_t \mathbf{I}),$
for \( t = 1, \ldots, T \), where \( \beta_{1:T} \) dictates the noise schedule. As 
$t$ approaches $T$, the data \( x_T \) approximates pure Gaussian noise \( \mathcal{N}(0, \mathbf{I}) \).

Generating samples from the noise-perturbed distribution $p(x_t)$ demands integrating over all steps, a computationally intensive process. However, the Gaussian choice facilitates generating any time-step $x_t$ using the closed-form expression:
\begin{equation}
x_t = \sqrt{\bar{\alpha}_t}x_0 + \sqrt{1 - \bar{\alpha}_t}\epsilon, \quad \epsilon \sim N(0, I),
\end{equation}
where $\alpha_t = 1 - \beta_t$ and $\bar{\alpha}_t = \prod_{s=1}^{t} \alpha_s$.

The diffusion model employs a variational Markov chain in the reverse process, parameterized as a time-conditioned denoising function $s(x, t; \theta)$, which is typically a neural network. The transition distribution can be expressed as $p_{\theta}(x_{t-1} | x_t) = \mathcal{N}\left(x_{t-1}; \frac{1}{\sqrt{1 - \beta_t}} (x_t + \beta_t s(x_t, t; \theta)), \beta_t I\right)$. The denoiser, tasked with minimizing a re-weighted variant of evidence lower bound (ELBO), fits its output to the score of a probability density $p(x_t)$:
\begin{equation}
\min_{\theta} ~ \mathbb{E}_{t,x_0,\epsilon} \left[ \left\| s(x_t, t; \theta) - \nabla_{x_t} \log p(x_t) \right\|_2^2 \right],
\end{equation}
where $\nabla_{x_t} \log p(x_t)$ is called the score function.


With a trained denoiser $s(x, t; \theta^*) \approx \nabla_{x_t} \log p(x_t)$, data generation is achieved by reversing the Markov chain:
\begin{equation}
x_{t-1} \leftarrow \frac{1}{\sqrt{1 - \beta_t}}(x_t + \beta_t s(x_t, t; \theta)) + \sqrt{\beta_t}\epsilon_t.
\end{equation}
This reverse process effectively traverses back through the noise addition, guided by the denoiser, to reveal the clean sample $x_0$.

\subsection{Low-rank adaptation}

The pursuit of efficiency in model adaptation often conflicts with the demand for high performance, typically requiring extensive parameter training. Low-rank adaptation~\cite{hu2021lora} emerges as a solution, leveraging the principles of low-rank matrix approximations to achieve parameter-efficient adaptation. This strategy focuses on subtle modifications within the neural network, capitalizing on the idea that meaningful changes can be efficiently encapsulated using a low-rank structure, reducing the necessity for extensive parameters.

Given a weight matrix $W\in \mathbb{R}^{d \times d}$, low-rank adaptation is applied through the introduction of \emph{low-rank adaptors}: 
\begin{equation}
\label{eq:lowrank}
W' = W + UV^T,
\end{equation} 
where $U \in \mathbb{R}^{d \times k}$ and $V \in \mathbb{R}^{d \times k}$, and $k \ll d$. The product $UV^T$ signifies a low-rank update to the original weight matrix, constraining the adaptation within a subspace and significantly cutting down the number of parameters involved. This efficient approach requires updates to only $O(kd)$ parameters, a stark reduction compared to the original $O(d^2)$, making the model adaptation notably more manageable and resource-efficient. The process strategically updates only the low-rank adaptors, leaving the backbone of the model frozen and preserving its original capabilities while enhancing performance. This method stands out in scenarios demanding quick adaptation and deployment, particularly where computational resources are limited.

\section{Details of Fourier High-frequency Loss}
\label{sec:rationale}
For Fourier high-frequency loss, we initially use the Discrete Fourier Transform (DFT) to transform the generated image \(I_{\text{generated}}\) from its spatial domain into the frequency domain. This transformation aids in the analysis and manipulation of the image's frequency components. The DFT is represented by the function \(F(u, v)\), which is a complex function representing the frequency domain of the image. The computation of DFT aims to break down the image into its sinusoidal components of varying frequencies:
\begin{equation}
F(u, v) = \sum_{m=0}^{M-1} \sum_{n=0}^{N-1} f(m, n) \cdot e^{-j2\pi \left(\frac{um}{M} + \frac{vn}{N}\right)}.
\end{equation} 

The next step is the application of a high-pass filter, denoted as \(M_{\text{high}}\). This filter is designed to isolate or emphasize the high-frequency components from \(F(u, v)\). It is represented as a matrix in the frequency domain, which effectively attenuates the low-frequency components and preserves the high-frequency components of the image spectrum.
The isolation is achieved through an element-wise multiplication of the high-pass filter \(M_{\text{high}}\) with the function \(F(u, v)\), resulting in \(F_{\text{high}}(u, v)\), which contains only high-frequency components. The operation is denoted by:
\begin{equation}
F_{\text{high}}(u, v) = F(u, v) \odot M_{\text{high}}(u, v).
\end{equation}

This outcome, \(F_{\text{high}}(u, v)\), highlights the details and intricacies within the image, which are often essential for improving the perceptual quality of generative models. These high-frequency components can later be transformed back to the spatial domain, depending on the requirements of the subsequent processing or analysis tasks. Following that, we use the Inverse Fourier Transform (IFT) for image reconstruction from the obtained high-frequency components. The process is mathematically represented as:

\begin{equation}
I_{\text{high-freq}} = \text{IFT}(F_{\text{high}}).
\end{equation}

The Fourier high-frequency loss function then aligns the high-frequency components from the one-step generated images with those from the original multi-step diffusion process based on Deep Image Structure and Texture
Similarity (DISTS)~\cite{ding2020image}:
\begin{equation}
L_{\text{Fourier}} = L_{\text{DISTS}}\left(I_{\text{freq}}^{\text{1-step}}, I_{\text{freq}}^{\text{multi}}\right).
\end{equation}

\section{More Experimental Results}\label{app_c}

\begin{table}[t] 
 \vspace{-0.1in}
 \begin{center}
 \begin{threeparttable} 
 \resizebox{0.35\textwidth}{!}{
 	\begin{tabular}{lcc}\toprule 
 Adaptation losses & FID-30k $\downarrow$ & CLIP $\uparrow$ \cr
 \midrule
 Stable Diffusion~\cite{rombach2022high} & 355.21	 	 & 0.11 \\
 ~~$\bullet~$L2 spatial loss & 131.20	 	& 0.16 \\
 
 ~~$\bullet~$LPIPS spatial loss & 52.30	 	& 0.23 \\
 ~~$\bullet~$DISTS spatial loss & 17.57 & 0.29\\ 
 ~~$\bullet~$HiPA loss (ours) & \textbf{15.89}		& \textbf{0.31} \\ 
 \bottomrule 
	\end{tabular}}
	 \end{threeparttable}
	 \end{center} 
 \vspace{-0.1in}
	\caption{One-step generation performance of various loss functions for adaptor training based on COCO 2014 validation set. For fast evaluation, all methods are trained for only one /epoch.} 
	\label{table_high_freq} 
 \vspace{-0.1in}
\end{table} 
 
\subsection{Effectiveness of loss function}
\label{sec:ablation_loss} 
In this appendix, we ablate the adaptation loss in HiPA for one-step text-guided image generation. Table~\ref{table_high_freq} strikingly illustrates the superior efficacy of our proposed loss. To be specific, the Stable Diffusion~\cite{rombach2022high} baseline yields an FID score of 355.21 and a CLIP score of 0.11. Using the proposed loss in HiPA results in a substantial enhancement, reducing the FID to a remarkable 15.89 and increasing the CLIP score to 0.31. This is a significant improvement compared to other adaptation losses like L2 spatial, LPIPS spatial, and DISTS spatial losses, which show varying degrees of performance enhancements. Notably, the DISTS spatial loss shows considerable effectiveness with an FID of 17.57 and a CLIP of 0.29, but it is the loss in HiPA that emerges as the most effective, achieving the best scores in both FID and CLIP metrics. This empirical evidence validates the effectiveness and superiority of the adaptation loss in HiPA.

\subsection{The choice of edge detection}
\label{sec:edge}

In HiPA, we leverage the Sobel operator to extract edge information, a critical component in enhancing the quality of one-step diffusion. However, one might wonder if other edge detection techniques could offer similar or improved results. In this appendix, we further compare the effectiveness of the Sobel operator with a Laplacian operator in the high-frequency edge loss. The Laplacian operator, another popular method for edge detection, operates by identifying regions of rapid intensity change.

As delineated in Table~\ref{table_high_freq_edge}, we observe a significant difference in the performance of HiPA when using these two operators. Specifically, the Sobel-based edge loss results in the best performance, with an FID score of 15.89 and a CLIP score of 0.31. This indicates its superior ability to capture the high-frequency details that contribute to the quality of one-step generated images.

\begin{figure}[t] 
 \centerline{\includegraphics[width=8.5cm]{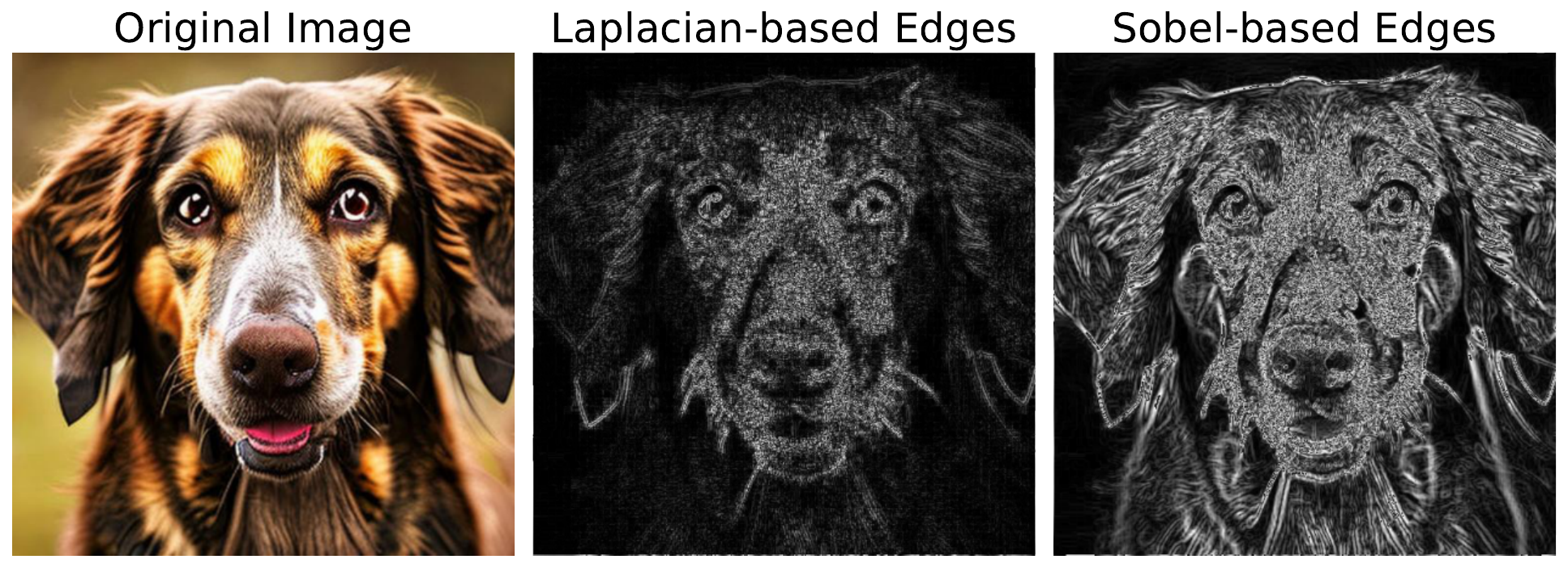}} 
 \vspace{-0.1in}
 \caption{Visualization of the detected edge information by Sobel operator and Laplacian operator.}\label{figure_edge} 
\end{figure}

Figure~\ref{figure_edge} visually substantiates this observation. The Sobel operator is seen to extract more detailed and rich edge information compared to the Laplacian operator. This enhanced detail retrieval is likely a key factor in the observed performance boost, as richer edge information contributes to more realistic and textually aligned image generation.

In contrast, the use of the Laplacian-based edge loss in HiPA results in slightly inferior performance. This might be attributed to the Laplacian operator's lesser sensitivity to fine detail compared to the Sobel operator.
In conclusion, our analysis demonstrates the effectiveness of the Sobel operator in the HiPA framework.

\begin{table}[t] 
 \vspace{-0.1in}
 \begin{center}
 \begin{threeparttable} 
 \resizebox{0.47\textwidth}{!}{
 	\begin{tabular}{lcc}\toprule 
 Methods & FID-30k $\downarrow$ & CLIP $\uparrow$ \cr
 \midrule
 HiPA w/o high-frequency edge loss & 17.02		& 0.30 \\
 HiPA w/ Laplacian-based edge loss & 18.29	 & 0.29 \\
 HiPA w/ Sobel-based edge loss & {15.89}		& {0.31} \\ 
 \bottomrule 
	\end{tabular}}
	 \end{threeparttable}
	 \end{center} 
 \vspace{-0.15in}
	\caption{Ablation on edge detection operators.} 
	\label{table_high_freq_edge} 
 \vspace{-0.1in}
\end{table}

\subsection{Are adaption layers equally important?} 
We next evaluate the impact of placing low-rank adaptors to various layers within the U-Net of Stable Diffusion. As shown in Table~\ref{table_various_layers}, when adaptors are applied to downscaling or upscaling blocks independently, there is a marked improvement compared to the baseline (no adaptors). Moreover, placing adaptors solely in middle blocks leads to only marginal improvement. The most significant enhancement is seen when adaptors are used across all layers, underscoring the synergistic effect of holistic adaptation. This comparison highlights the varying influence of low-rank layers in different sections of the model, with upscaling layers proving to be more influential in enhancing image quality for one-step generation.

\begin{table}[t] 
 \vspace{-0.15in}
 \begin{center}
 \begin{threeparttable} 
 \resizebox{0.47\textwidth}{!}{
 	\begin{tabular}{ccccccc}\toprule 
 \multicolumn{3}{c}{Position of HiPA adaptors} && \multirow{2}{*}{FID-30k $\downarrow$} & \multirow{2}{*}{IS $\uparrow$} & \multirow{2}{*}{CLIP $\uparrow$} \\\cmidrule{1-3}
 Down blocks & Mid blocks & Up blocks & & & & \cr
 \midrule
 & & & & 355.21	& 1.97 & 0.11 \\
 \cmark & & & & 40.09 	& 19.18 & 0.26 \\
 &\cmark & & & 295.01 & 2.28 & 0.06 \\
 & & \cmark& & 19.97 & 24.75 & 0.28 \\
 \cmark & \cmark & \cmark && \textbf{15.89} & \textbf{27.29} & \textbf{0.31} \\ 
 \bottomrule 
	\end{tabular}}
	 \end{threeparttable}
	 \end{center} 
 \vspace{-0.15in}
	\caption{Impact of low-rank adaptors across different layers on one-step generation of HiPA based on MS-COCO 2014. All methods are trained for only one epoch.}
	\label{table_various_layers} 
 \vspace{-0.15in}
\end{table} 

\subsection{Ablation studies on the adaptor ranks} 
In our method, we default the rank of HiPA adaptors to 16. This appendix further analyzes the influence of the rank. As shown in Table~\ref{table_ablation}, we can draw three main observations.

\textbf{Quality enhancement with rank increase}: There is a clear trend that as the rank of adaptors increases from 4 to 16, both the FID-30k and IS scores improve. This implies that higher ranks, despite requiring more parameters (from 0.8 million to 3.3 million), effectively boost the quality and diversity of generated images.

\textbf{Diminishing returns at higher ranks}: However, upon further increasing the rank to 32, the improvement in FID-30k becomes marginal, and IS score slightly decreases, even as the number of parameters grows to 6.6 million. This suggests a diminishing return at higher ranks, indicating an optimal rank beyond which additional parameters do not equate to significant quality gains.

\textbf{Balancing efficiency and performance}: The results imply that a balance must be struck between model complexity and generation performance. A rank of 16 appears to offer a good compromise, enhancing quality without an excessive increase in parameters. This balance is crucial for real applications where computational efficiency is important.

\begin{table}[t] 
 \vspace{-0.15in}
 \begin{center}
 \begin{threeparttable} 
 \resizebox{0.45\textwidth}{!}{
 	\begin{tabular}{ccccc}\toprule 
 Rank & FID-30k $\downarrow$ & IS $\uparrow$ & CLIP $\uparrow$ & \# parameters \cr
 \midrule
 4 & 17.09 	& 26.45 & 0.29 & 0.8 million	 \\
 8 & 16.73 & 26.77 & 0.30 & 1.7 million \\
 16 & 15.89 & 27.29 &	 0.31 &	 3.3 million \\
 32 & 15.86& 27.16& 0.31 & 6.6 million \\ 
 \bottomrule 
	\end{tabular}}
	 \end{threeparttable}
	 \end{center} 
 \vspace{-0.15in}
	\caption{Impact of the adaptor rank on HiPA for one-step diffusion on MS-COCO 2014 validation set. The guidance scale is 2.}
	\label{table_ablation} 
\end{table}

\subsection{Ablation studies on learning rates} 
In this work, we default the learning rate to 1e-4.  Table~\ref{table_ablation_lr} further evaluates the impact of learning rates on the performance of HiPA in one-step diffusion. Optimal performance is achieved with a learning rate of 1e-4, where the model attains the lowest FID-30k score of 15.89 and the highest IS and CLIP scores, indicating superior image quality, diversity, and textual alignment. Both higher (1e-2 and 1e-3) and lower (1e-5 and 1e-6) learning rates result in significantly poorer performance, underscoring the importance of a balanced learning rate (1e-4) for efficient training and optimal model output.

\begin{table}[t] 
 \begin{center}
 \begin{threeparttable} 
 \resizebox{0.3\textwidth}{!}{
 	\begin{tabular}{cccc}\toprule 
 LR & FID-30k $\downarrow$ & IS $\uparrow$ & CLIP $\uparrow$ \cr
 \midrule
 1e-2 & 467.2	& 1.0 & 0.17 	 \\
 1e-3 & 19.28 	& 24.22 & 0.29 	 \\
 1e-4 & 15.89 & 27.29 &	 0.31 \\
 1e-5 & 25.36 & 23.28 &	0.30 \\
 1e-6 & 137.1& 8.35& 0.20 \\ 
 \bottomrule 
	\end{tabular}}
	 \end{threeparttable}
	 \end{center} 
 \vspace{-0.15in}
	\caption{Impact of the learning rate (LR) on HiPA for one-step diffusion on MS-COCO 2014 validation set. }
	\label{table_ablation_lr} 
\end{table}

\subsection{Ablation studies on batch sizes} 
In this work, we default the batch size to 8.  
The results in Table~\ref{table_ablation_bs} show the influence of batch sizes on HiPA in one-step diffusion. An interesting trend is observed: smaller batch sizes (1 and 2) lead to slightly better FID-30k scores, indicating a marginal increase in image quality. However, as batch size increases (4, 8, and 16), there's a gradual decrease in performance, as evidenced by higher FID-30k and lower IS scores. This suggests that while larger batch sizes may aid in computational efficiency, they slightly compromise the model's ability to generate diverse and high-quality images. Nevertheless, the performance remains consistently high across all batch sizes, with CLIP scores remaining stable, indicating that textual alignment is relatively unaffected by changes in batch size.

\begin{table}[t] 
 \begin{center} 
 \begin{threeparttable} 
 \resizebox{0.3\textwidth}{!}{
 	\begin{tabular}{cccc}\toprule 
 BS & FID-30k $\downarrow$ & IS $\uparrow$ & CLIP $\uparrow$ \cr
 \midrule
 1 & 14.10	& 28.28 & 0.30 	 \\
 2 & 14.49	& 27.97 & 0.31	 \\
 4 & 15.67 & 37.35 &	 0.31 \\
 8 & 15.89 & 27.29 &	 0.31 \\
 16 & 16.02 & 26.06& 0.30 \\ 
 \bottomrule 
	\end{tabular}}
	 \end{threeparttable}
	 \end{center} 
 \vspace{-0.15in}
	\caption{Impact of the batch sizes (BS) on HiPA for one-step diffusion on MS-COCO 2014 validation set. }
	\label{table_ablation_bs}  
 \vspace{-0.15in}
\end{table}

\subsection{Ablation studies on cutoff values} 
In this work, we default the cutoff value in high-frequency Fourier loss to 5. Table~\ref{table_ablation_cutoff} further ablates the effect of the cutoff value on HiPA in one-step diffusion. The table reveals three primary observations:

\textbf{Optimal cutoff value for performance}: A cutoff value of 5 yields the most balanced outcome in terms of FID-30k and Inception Score (IS), indicating optimal performance for image quality and diversity. This suggests that a cutoff value of 5 effectively balances high-frequency detail extraction without overemphasizing noise or losing essential image details.

\textbf{Variations in cutoff and image quality}: As the cutoff value changes from 5 to 10, there is a slight increase in FID-30k and a decrease in IS, implying a reduction in image quality and diversity. This might be due to higher cutoff values excessively filtering out important high-frequency information, leading to a loss of detail and texture in the generated images.

\textbf{Consistency in textual alignment}: Across different cutoff values, the CLIP scores remain relatively stable, indicating that the alignment with textual descriptions is not significantly impacted by the variation in the cutoff value. This underscores the robustness of HiPA in maintaining textual consistency irrespective of the changes in high-frequency detail capture.

In conclusion, we suggest that the choice of the cutoff value is crucial in fine-tuning the balance between detail capture and noise avoidance, with a value of 5 presenting an effective compromise for one-step diffusion in HiPA.

\begin{table}[t] 
 \begin{center}
 \begin{threeparttable} 
 \resizebox{0.38\textwidth}{!}{
 	\begin{tabular}{cccc}\toprule 
 Cutoff & FID-30k $\downarrow$ & IS $\uparrow$ & CLIP $\uparrow$ \cr
 \midrule
 1 & 15.74	& 26.86 & 0.31	 \\
 5 & 15.89 & 27.29 &	 0.31 \\
 10 & 16.13 & 27.17 & 0.30 \\ 
 \bottomrule 
	\end{tabular}}
	 \end{threeparttable}
	 \end{center} 
 \vspace{-0.15in}
	\caption{Impact of the cutoff value on HiPA for one-step diffusion on MS-COCO 2014 validation set. The guidance scale is 2.}
 \label{table_ablation_cutoff} 
\end{table}

\subsection{Ablation studies on smoothness regularizer} 
Table~\ref{table_ablation_smoothness} shows the significance of a smoothness regularizer in HiPA for one-step diffusion:

\textbf{Quality improvement}: Adding a smoothness regularizer enhances image quality, as reflected by the improved FID-30k score. This indicates better visual appeal and reduced artifacts in generated images.

\textbf{Increased diversity and alignment}: The smoothness regularizer not only betters image quality but also boosts the Inception Score and the CLIP score. This implies a richer diversity in the generated images and stronger alignment with text prompts.

In essence, the smoothness regularizer is beneficial to HiPA, crucially enhancing the overall quality, diversity, and textual alignment in one-step diffusion models.

\begin{table}[t] 
 \begin{center} 
 \begin{threeparttable} 
 \resizebox{0.45\textwidth}{!}{
 	\begin{tabular}{lccc}\toprule 
 Methods & FID-30k $\downarrow$ & IS $\uparrow$ & CLIP $\uparrow$ \cr
 \midrule
 HiPA w/o smoothness & 17.73	& 26.73	& 0.30 \\ 
 HiPA & {15.89} & 27.29		& {0.31} \\ 
 \bottomrule 
	\end{tabular}}
	 \end{threeparttable}
	 \end{center} 
 \vspace{-0.15in}
	\caption{Ablation on smoothness regularizer.}  
	\label{table_ablation_smoothness} 
\end{table}

\section{More Visualizations}\label{app_d}

\subsection{More results on one-step image generation} 

\begin{figure*} 
 \vspace{-0.1in}
 \centering 
 \begin{minipage}{.47\textwidth}
 \centering
 \includegraphics[width=\linewidth]{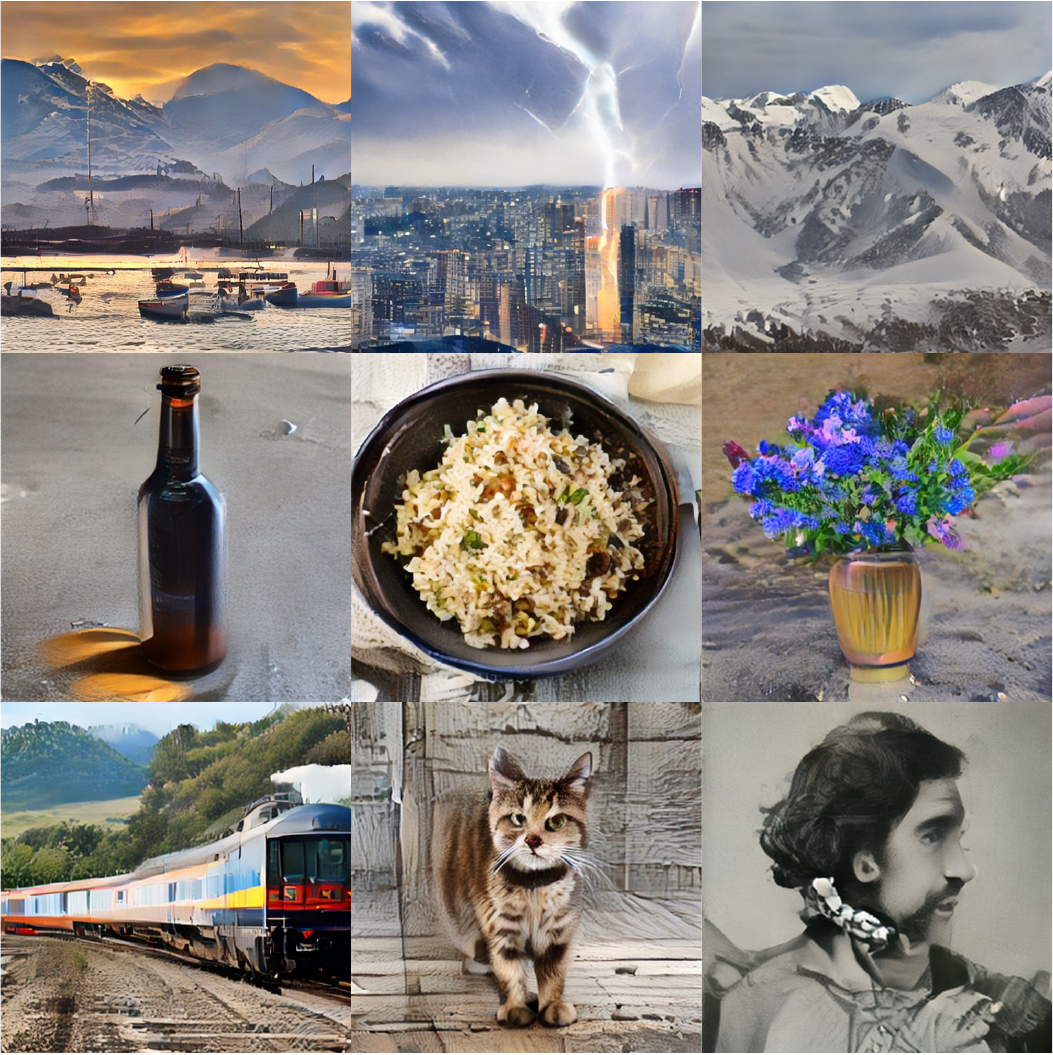}
 \vspace{-0.25in}
 \caption*{(a) One-step generation by HiPA-adapted SD (ours)} 
 \end{minipage} 
 \hfill 
 \begin{minipage}{.47\textwidth}
 \centering
 \includegraphics[width=\linewidth]{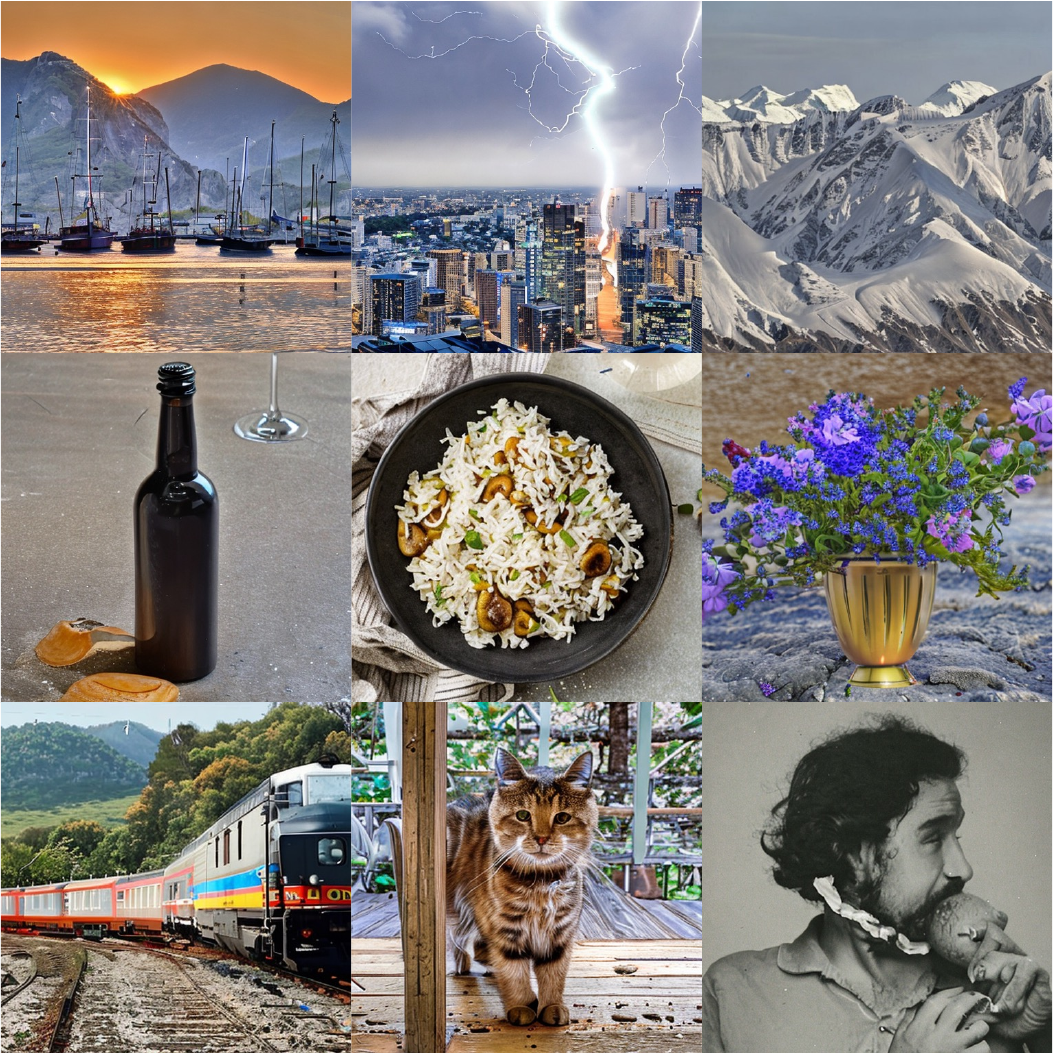}
 \vspace{-0.25in}
 \caption*{(b) Multi-step generation by SD} 
 \end{minipage}
 \vspace{-0.1in}
 \caption{More visualization results. (a) More visualizations of one-step text-guided image generation on MS-COCO 2014 by our HiPA-adapted model. (b) visualization results of multi-step text-guided image generation by Stable Diffusion (SD) based on DPM-solver.}
\label{visualization_qualitative_supp} 
\end{figure*}

We provide more visual results of one-step text-guided image generation by HiPA in Figure~\ref{visualization_qualitative_supp}(a). These results further demonstrate the effectiveness of HiPA for one-step text-to-image diffusion generation.
 
Despite HiPA's effectiveness, some artifacts are still observed in the one-step generated images (see Figure~\ref{visualization_qualitative_supp}(a)). 
However, many of these issues may come from the original multi-step SD models, such as distorted faces and fingers (cf. Figure~\ref{visualization_qualitative_supp}(b)). Future improvements could explore adapting more sophisticated diffusion models, like SD-XL~\cite{podell2023sdxl} and DALL-E3~\cite{shi2020improving}, subject to the availability of GPUs with higher memory capacities.

Moreover, Figure~\ref{visualization_qualitative_supp} also reflects that HiPA can serve as a viable solution for rapid preliminary image generation and review. Users can utilize it for initial drafts and select their preferred drafts. Following that, users can revert to the original SD model, sans HiPA adaptors, for refining and producing higher-quality final images, effectively balancing speed with details.

 \subsection{More results on one-step image editing} 
\begin{figure*}[t] 
	\centering
	\includegraphics[width=1\linewidth]{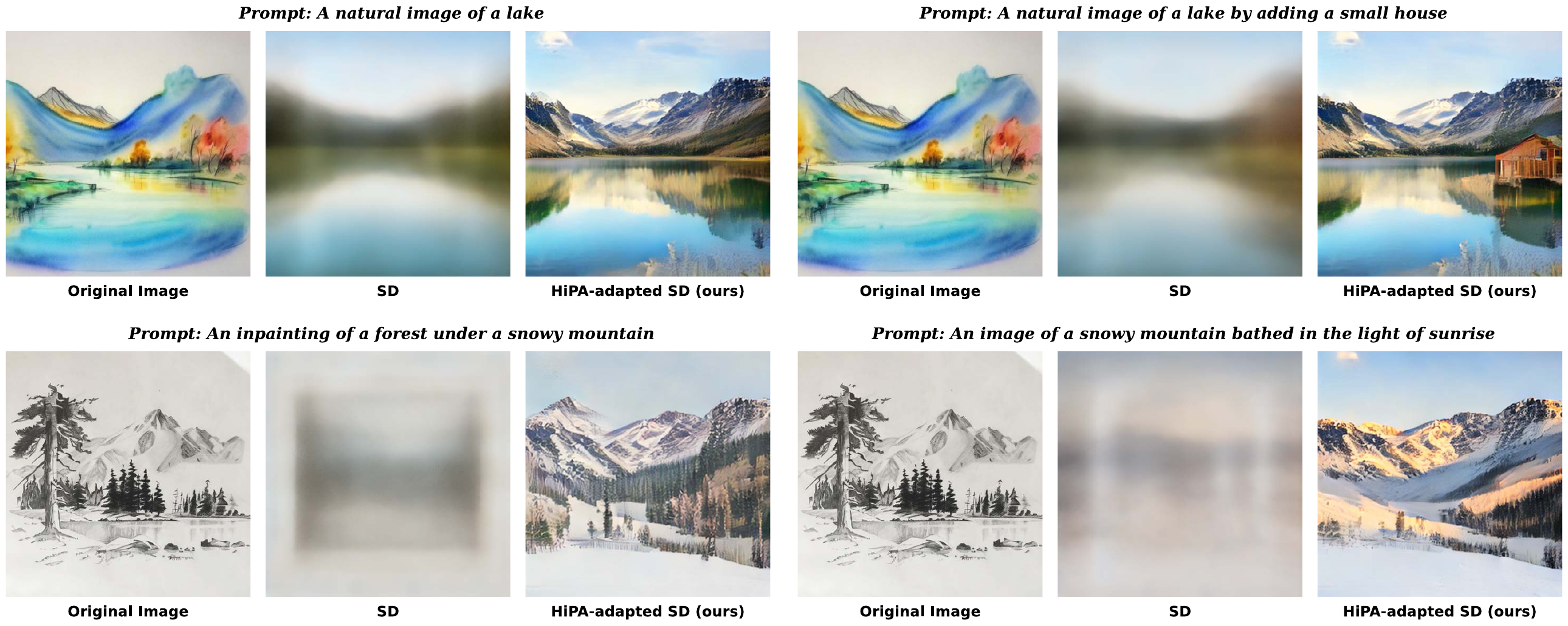}
 \vspace{-0.2in}
	\caption{More visualizations of one-step text-guided image editing by Stable Diffusion (SD) and our HiPA-adapted SD model.} 
	\label{visualization_image2image_supp} 
\end{figure*}

Further visual examples of one-step text-guided image editing using HiPA are presented in Figure~\ref{visualization_image2image_supp}. These examples highlight HiPA's capability for rapid and effective text-guided image modifications in real-world scenarios.

\subsection{More results on one-step image inpainting} 

\begin{figure*}[t]
	\centering
	\includegraphics[width=1\linewidth]{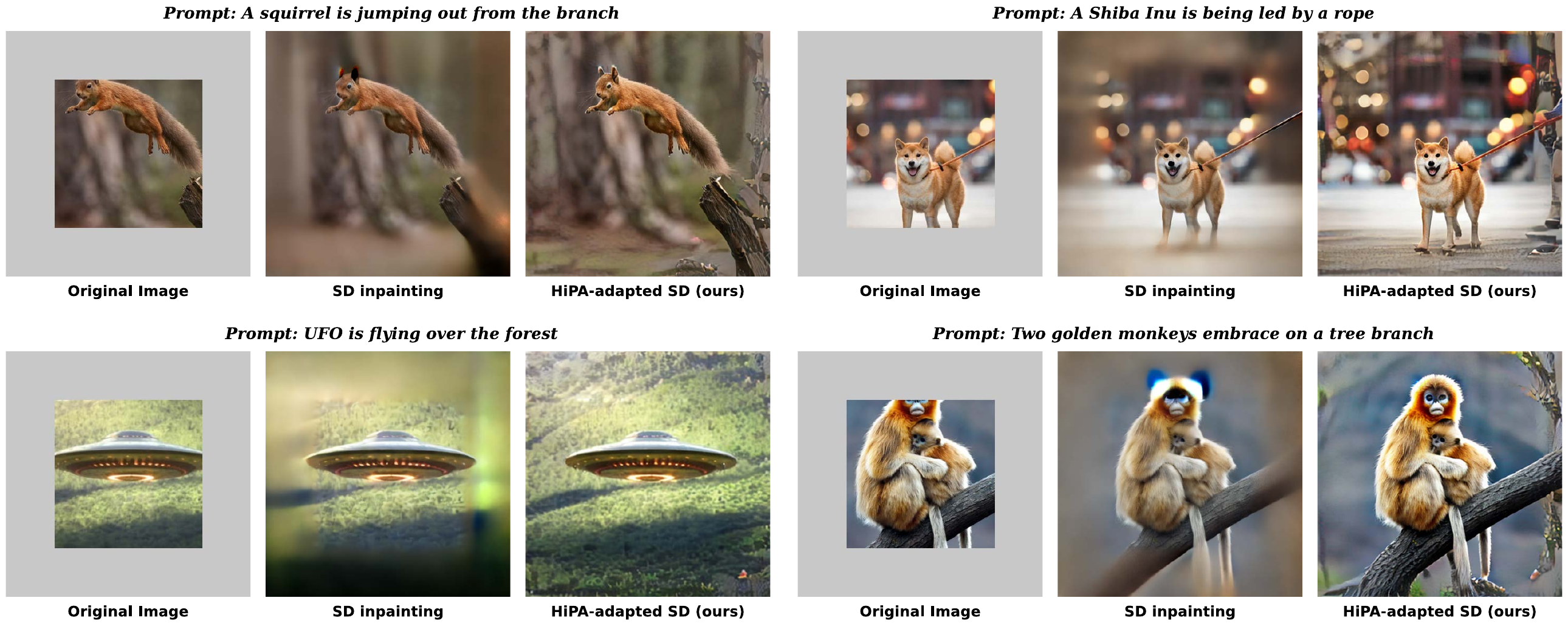}
 \vspace{-0.2in}
	\caption{More visualization examples of one-step text-guided image inpainting by Stable Diffusion (SD) inpainting model and our HiPA-adapted SD inpainting model.}
	\label{visualization_inpainting_supp} 
 
\end{figure*}

Additional visualizations of one-step text-guided image inpainting are displayed in Figure~\ref{visualization_inpainting_supp}. These examples underscore HiPA's ability to swiftly and effectively conduct text-guided image inpainting in practical applications.

\subsection{More results on one-step super-resolution}
\begin{figure*}[t] 
\centering
 \includegraphics[width=1\linewidth]{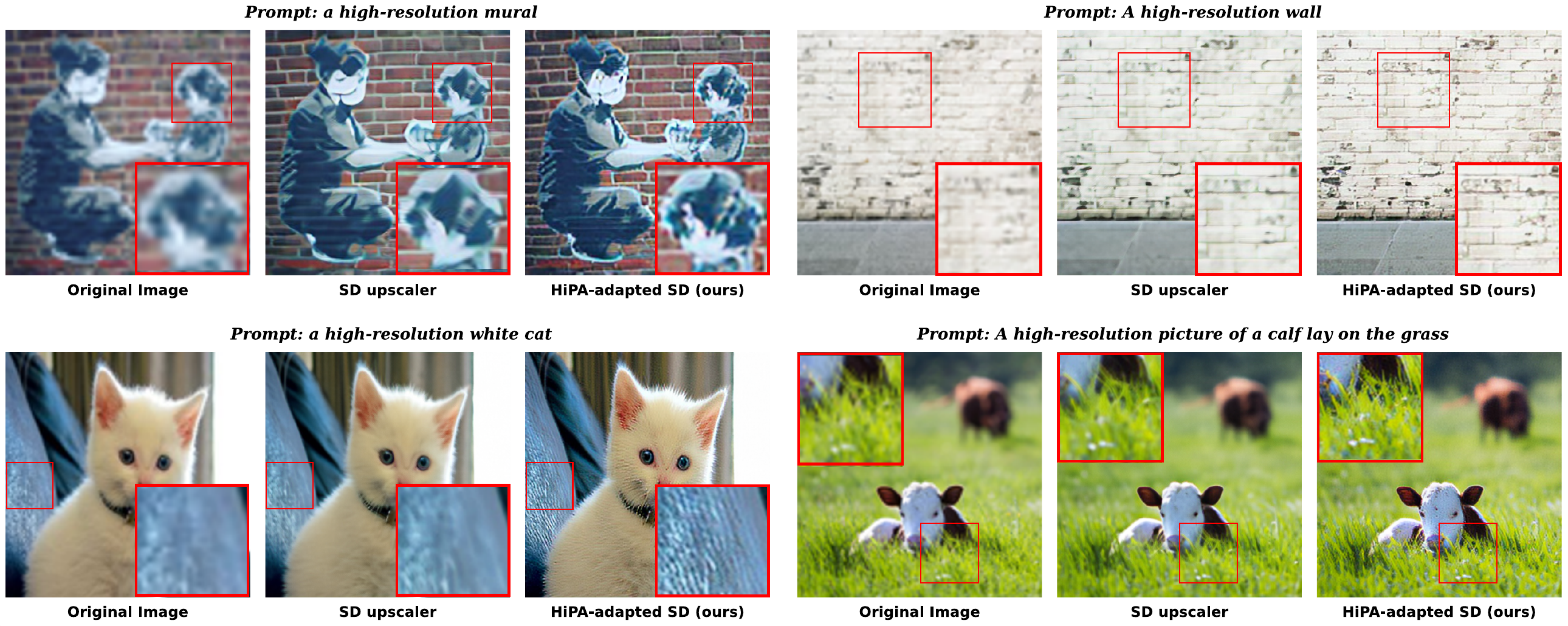}
 \vspace{-0.2in}
 \caption{More visualization examples of one-step text-guided image super-resolution by Stable Diffusion (SD) upscaler model and our HiPA-adapted upscaler model.} 
\label{visualization_sr_supp} 
\end{figure*}

We provide more visualized results on one-step text-guided image super-resolution tasks in Figure~\ref{visualization_sr_supp}. The results highlight the effectiveness of HiPA to perform fast text-guided image super-resolution in real applications.

\end{document}